# Cross-Disciplinary Taxonomy and Modeling of Misunderstanding Generation, Amplification, and Detection, from Pragmatics to AI Agents

*Babak Abbaschian*
*University of Louisville*
*ORCID: https://orcid.org/0000-0003-4876-9372*

## Abstract

Detection of misunderstanding is an urgent problem to solve because communication has moved away from real-time, in-person interaction and is increasingly handled by AI-mediated channels. This shift cuts communicators off from the resources repair depends on faster than new means of detection are being built. In this paper we analyse misunderstanding as a layered process in which a divergence is generated, may then be amplified, and is either detected and repaired or left to persist unnoticed. Consolidating accounts from nine fields of research that do not ordinarily cite one another, we identify eleven exact failure modes and show that each operates at a specific point in a communicative process rather than anywhere within it. Those points give eight analytical layers, derived from the literature rather than adopted from an existing model. Eight of the mechanisms primarily generate a divergence, two primarily amplify one already present, and one governs whether a divergence is detected and repaired. We model the eight layers formally, extending information and communication theory from the transmission of signals to the reconstruction of meaning, and we supply a source-by-source evidence matrix that makes every rating auditable, a coding manual, and nine analysed dialogue cases. No prior classification of misunderstanding both locates mechanisms at points in the process and types them by function.

## 1. Introduction

### 1.1 The Problem

Human two-party communication is remarkably robust. People routinely coordinate their actions, maintain their relationships, and accomplish shared goals, despite substantial differences in background knowledge, experience, and language. This everyday success demonstrates the efficiency of ordinary human communication. However, the same communicative machinery remains vulnerable to a particular form of failure, in which the meaning a listener reconstructs diverges from the meaning the speaker intended. That divergence is most consequential when neither party recognises that it has occurred, which is the condition we refer to as an undetected misunderstanding. That condition is the one this paper is organised around, and it is becoming both more likely and harder to notice as AI agents take up position between communicating parties.

Research on this form of communicative failure is highly fragmented. Linguists examine it as pragmatic failure. Conversation analysts study it as a trouble source in the moment-to-moment organization of talk. Cognitive psychologists trace it to egocentric interpretation and to failures of perspective-taking. Organizational scholars treat it as a breakdown in coordination. Electrical and communication engineers locate it in encoding failures, channel noise, and information loss. Researchers in healthcare and law document its consequences in patient harm and in litigation (Charrow & Charrow, 1979; Haig et al., 2006; Joint Commission, 2017; Solan, 2004). However, these fields rarely share terminology, concepts, or a common analytical framework, and a mechanism identified in one is frequently reinvented under another name in the next.

The result is a literature that is deep within individual fields but poorly integrated across them. What is missing is a mechanism-level map, by which we mean an account of the specific processes through which misunderstandings are generated, amplified, detected, or allowed to persist. Such a map must be expressed in terms that researchers across fields can use. It must also be applicable at more than one level of analysis.

The stakes of that gap are rising rather than holding steady. For as long as people have talked to one another, misunderstanding has been endemic and yet survivable, because the means of correcting it are continuous, cheap, and largely automatic. Repair is initiated roughly once every 1.4 minutes, and it recurs in similar forms across widely separated languages (Dingemanse et al., 2015). What has changed is not the rate at which misunderstandings are generated. It is the rate at which they are caught.

Every step away from co-present conversation removes some of the resources that repair depends on. Lean media supply fewer cues for resolving an ambiguous message (Daft & Lengel, 1984, 1986), each medium withdraws a particular grounding resource (Clark & Brennan, 1991), and a system interposed between participants can remove the mutual monitoring through which trouble used to become visible (Heath & Luff, 2000). However, the most consequential step is also the most recent one. When an Agentic AI system, modifies, augments, or generates messages on a communicator's behalf (Hancock et al., 2020), a link enters the chain that has no capacity to notice its own misreading and no social reason to hesitate. People are measurably less likely to initiate repair when they know the party they are addressing is an artificial agent, and they invest the most effort in establishing common ground under precisely the conditions that most resemble face-to-face human interaction (Corti & Gillespie, 2016).

Consider a sequence of four exchanges, each of which is an ordinary two-party episode of the kind this paper analyses, and each of which has an AI agent on at least one side. A head of state instructs an AI assistant system to prepare correspondence to a defence ministry, without specifying whether the action sought is preparatory or executive, because the distinction is obvious to the speaker and therefore goes unstated. The AI assistant system resolves that underspecification towards the more consequential reading and composes accordingly, and nothing in the exchange surfaces the choice it has made. The receiving ministry's AI assistant system parses the correspondence as a directive rather than as a query, because in inter-ministerial correspondence the two forms are nearly identical on the surface. The minister reads the resulting summary and takes the absence of hedging as a signal of urgency, rather than as an artifact of summarisation, and officials are known to defer to algorithmic output in exactly this way(Alon-Barkat & Busuioc, 2023; Passi & Vorvoreanu, 2022).

No exotic failure occurs anywhere in that sequence. Each hop instantiates a mechanism catalogued in this paper, being an underspecified intention, an inference drawn from different background assumptions, a misidentified communicative act, and an overestimation of what is shared. What makes the outcome different from any one of them is that no hop contains an opportunity for repair, so a divergence introduced at the first exchange is still present, unexamined and by now authoritative, at the fourth. Comparable compounding has been documented empirically in chains of AI agents, where communication breakdown between agents is one of three top-level categories of system failure (Cemri et al., 2025). It is already late to be doing this work. The conditions that have always made misunderstanding survivable are being removed faster than the means of detecting divergence are being built, and a mechanism-level account of how divergences arise and how they are caught is the precondition for building detection back into channels that no longer supply it.

## 1.2 Nature and Scope of Study

We present an integrative map of misunderstanding mechanisms, developed through a cross-disciplinary conceptual synthesis. This is a conceptual article of the typology type, developed through theory synthesis (Jaakkola, 2020), and an integrative review in the established sense (Cronin & George, 2023). Its purpose is to combine concepts and evidence from multiple fields in order to construct a new theoretical framework. The standard governing such work is that the dimensions of a typology, its categories, and the grounds for both are made explicit, and that the result is specified well enough to be tested empirically (Doty & Glick, 1994). We hold the synthesis to that standard and exceed it in two respects. Appendix A records, for every mechanism, each supporting source, the research body it belongs to, the type of evidence it supplies, and the basis for its inclusion, so that any rating can be disputed source by source. Appendix B supplies observable indicators, exclusion criteria, and a nearest-neighbour test for each mechanism, so that others can apply the taxonomy and measure their agreement. This study is not designed as a scoping review (Arksey & O'Malley, 2005). Its objective is theory synthesis and typology construction rather than the comprehensive mapping of a bounded evidence corpus, and we therefore do not claim compliance with PRISMA-ScR (Tricco et al., 2018).

The procedures we used to identify, select, and analyse the literature are presented alongside the literature review itself. The procedures we used to assess convergence and develop the taxonomy are presented alongside the taxonomy. The limitations arising from this design are addressed in Section 6.6.

The synthesis asks four questions. First, which mechanisms of misunderstanding recur across the nine fields of research. Second, how strong the cross-disciplinary support for each mechanism is. Third, whether the mechanisms can be organized according to whether they generate a divergence, amplify an existing divergence, or govern its detection and repair. Fourth, whether the resulting taxonomy can be applied to real and constructed dialogue cases at the level of specific mechanisms.

The scenario above is a chain rather than a dyad, and the analysis in this paper is confined to the dyadic link. We treat a chain as a composition of two-party episodes, on the grounds that a mechanism operating across a chain must first operate across one of its links. The propagation and compounding behaviour of chains is outside our scope, and it is the reason for establishing the dyadic case first.

## 1.3 Contributions

In this section, we enumerate the five contributions this paper makes to the study of misunderstanding. First, we bring together nine fields of research that do not ordinarily cite one another, and we extract from each the aforementioned vocabulary of misunderstanding coupled with the failure modes it names. Consolidating the accounts that describe the same process under different names leaves eleven distinct mechanisms, each written as an exact failure mode rather than as a general risk factor. The eleven cover communication between people and between people and machines under one scheme.

Second, we show that these mechanisms do not operate anywhere in a communicative episode but at particular points within it, and that the eleven fall between only eight such points. Those eight analytical layers are derived from the literature rather than adopted from an existing model, and we set out what happens at each and which mechanisms operate there.

Third, we give the account a formal statement. Section 3 defines a misunderstanding as a divergence that exceeds a materiality threshold, and Section 4.2 writes the eight layers as a chain of conditional distributions carrying an intended meaning to a reconstructed one, with reconstruction treated as inference under loss and two structural properties following from that factorization. The formal treatment is what allows a mechanism to be stated as a perturbation of one conditional distribution, rather than as a named category alone.

Fourth, we type each mechanism by what it does, as generating a divergence, amplifying one already present, or governing whether one is detected. Eight primarily generate, two primarily amplify, and one governs detection and repair. We refer to this as generate, amplify, and detect, abbreviated as GAD. The typing was not imported and then applied. It emerged from the mechanisms themselves, and it then proved to reproduce a distinction that conversation analysis, human-error theory, and root-cause analysis had each reached on their own evidence. We treat that convergence, however, as a check on the taxonomy rather than as a borrowing.

Fifth, we make the result checkable. Appendix A gives the source-by-source evidence behind every convergence rating, Appendix B gives a coding manual for applying the taxonomy to cases, and Section 5 applies it to nine dialogue cases. We show how a misunderstanding can be analysed in terms of specific primary and contributing mechanisms, rather than described through a broad label such as "poor communication." Against the classifications of misunderstanding that already exist, three of these contributions have no precedent that we could find. Each of those classifications stays within a single field, none of them gives its mechanisms a formal statement, and we have found none that both locates a mechanism at a point in the communicative process and independently types it by the function it performs. Section 2.10 sets out what exists and where each of them stops.

### 1.4 Organization of the Paper

The rest of the paper is organized as follows. In Section 2, we review the literature across nine fields of research and describe how the sources were identified, selected, and analysed. In Section 3, we derive a working definition of misunderstanding and establish its conceptual boundaries and its detection and resolution states. In Section 4, we present the taxonomy: eleven mechanisms, the eight layers at which they operate, three functional roles, and a convergence assessment of the evidence behind each mechanism. In Section 5, we apply the taxonomy to nine dialogue cases. Section 6 discusses the principal findings, what the nine cases show, the implications for communication mediated by AI agents, and the gaps and limitations of the synthesis, and Section 7 concludes.

## 2. Cross-Disciplinary Literature Review

In this section, we will examine the literature and research from various fields addressing misunderstandings from different perspectives. Our purpose is to establish the conceptual basis for the definition and the taxonomy that follows. Because these literatures differ in terminology, assumptions, methods, and units of analysis, we review each within its own conceptual context before comparing recurring mechanisms across fields.

To do so, we have grouped the literature into their own respective fields, resulting in nine groups of scientific fields. These groupings are contextual analytical rather than pure disciplinary, developed for this review rather than drawn from an established taxonomy of fields. We included each grouping because it offers a distinct account of how a misunderstanding may arise, develop, become amplified, or remain undetected. However, the boundaries between them are ours, and a different partition of the same literature would be defensible.

We have included foundational theory, experimental studies, conversation-analytic research, corpus studies, and applied evidence. We retained a source when it identified, tested, or operationalized a specific mechanism, or when it helped to distinguish misunderstanding from a related phenomenon. For each source, we have recorded the concepts and terminology it uses, and the point in the communication process at which its mechanism operates. We also have recorded the form of evidence presented, and whether a comparable mechanism appears in other fields.

### 2.1 Electrical and Communication Engineering: Information and Communication Theory

The foundational treatment of this problem is a mathematical theory of signal transmission, concerned with how accurately a signal sent from one point can be decoded, or reconstructed at another (Shannon, 1948). Weaver's later chapter extended that framework by distinguishing three problems (Shannon & Weaver, 1949). The technical problem asks how accurately the symbols of communication can be transmitted. The semantic problem asks how precisely the transmitted symbols convey the intended meaning.

The effectiveness problem asks how far the received meaning affects conduct in the way intended. However, the two contributions should not be treated as identical, and the distinction matters for what follows. Shannon's formal theory concerns the encoding, transmission and decoding of signals, whereas Weaver uses it as a basis for broader questions of meaning and effect.

Shannon's model represents communication as a source, transmitter, channel, receiver, and destination. Let $X$ denote the transmitted message and $Y$ the received signal. The entropy of the source is:

$$H(X) = -\sum_{x} p\,(x)\log_2 p(x),$$

which measures the average uncertainty associated with the possible messages produced by the source.

For a noisy channel, Shannon defines the conditional entropy:

$$H(X \mid Y) = -\sum_{x,y} p\,(x,y)\log_2 p(x \mid y).$$

This quantity measures the remaining uncertainty about what was transmitted after the received signal is known. Shannon calls it equivocation. The information conveyed through the channel is therefore:

$$I(X;Y) = H(X) - H(X \mid Y),$$

and the capacity of a noisy channel is:

$$C = \max_{p(x)} I(X;Y).$$

When $H(X \mid Y) = 0$, the transmitted message can be reconstructed from the received signal without residual uncertainty. As $H(X \mid Y)$ increases, the received signal leaves greater uncertainty about what was sent.

This formulation identifies two communication failures that are relevant to the present synthesis. The first is a loss of fidelity in encoding or decoding, and the second is distortion introduced by the channel. It also provides a formal distinction between what is transmitted, what is received, and how much uncertainty remains between the two.

However, Shannon's theory does not determine whether the receiver reconstructed the meaning that the sender intended. Two participants may receive the same signal with complete accuracy while assigning different meanings to it. The reason is that Shannon's framework assumes encoding and decoding rules that are known and fixed within the communication system. The encoding and decoding performed by communicating people are private, world-model dependent, and free to drift apart. Shannon's equivocation therefore supplies the structural basis for modelling divergence between input and reconstruction. However, extending that structure from signals to meanings requires an additional semantic formulation, which we develop in Section 3.

Weaver's second and third problems are now called semantic and task-oriented communication, where a system is designed to convey what a message is for, rather than which symbols it contains, and is evaluated on the task the receiver performs rather than on the bits it recovers (Gunduz et al., 2023). That work is close to our synthesis in its object and far from it in its assumptions. Semantic communication is built on the premise that transmitter and receiver share, or can be jointly trained to share, a model of the context in which the exchange takes place. The problem addressed here is what happens when they cannot. Between people the encoding and decoding models are private, acquired separately, and free to drift without either party noticing, and there is no design stage at which they can be aligned, they can be learned implicitly by reinforcing, but also can be mal-learned by the same mechanism, as there is no explicit labeling or reward system. That literature therefore optimises the conveyance of meaning under conditions where semantic mismatch has been engineered away, whereas the mechanisms reviewed here describe what happens when it has not.

The distance between the two narrows as one endpoint becomes a machine. When a message passes through an AI agent, one side of the exchange does hold a designed and inspectable model, and the two literatures begin to describe the same episode from opposite ends. In Weaver's terms, this paper works at the second and third problems rather than the first. The divergence it defines is semantic, and the threshold that makes a divergence material is defined by effect on conduct.

Our analysis places the failure this literature describes at the stage of formulation, where an intended meaning is put into words or other signals. A meaning may be expressed in a form that permits a reading other than the one intended, and we carry that forward as Encoding Ambiguity.

## 2.2 Linguistics and Philosophy of Language: Pragmatic Inference

Conversation operates through a Cooperative Principle together with four maxims (Grice, 1975). In this formulation, a speaker is expected to provide as much information as is needed (Quantity), to be truthful (Quality), to be relevant (Relation), and to be clear (Manner). Listeners draw on these expectations to infer meanings that go beyond the literal words. However, the same expectations generate misunderstanding when the listener derives an implicature different from the one intended, because the two parties reason from different background assumptions.

In addition to that, the Relevance Theory proposes that a listener accepts the first accessible interpretation that satisfies their expectations of relevance (Sperber & Wilson, 1995). When the listener's contextual assumptions differ from the speaker's, that same economy may deliver an interpretation the speaker never intended.

The Graded Salience Hypothesis adds that people prioritise meanings according to familiarity, prior knowledge, and conventional use (Giora, 1997, 2003). The meaning most salient to the listener may therefore differ from the one the speaker intended, without either party having reasoned carelessly.

While these articles explain the failure differently, they converge on a single stage, the listener's inference and interpretation, where an implied meaning other than the one intended is recovered. So we consolidate that convergence as Pragmatic Inference Failure.

## 2.3 Speech-Act Theory

Speech-act theory distinguishes three aspects of any utterance (Austin, 1962), which is a source of several failure mechanisms in our taxonomy. The locutionary act is what is literally said, the illocutionary act is what the speaker is doing in saying it, and the perlocutionary effect is what the utterance brings about in the listener (Searle, 1969). Requesting, warning, promising, criticising, and joking are all illocutionary acts. Misunderstanding occurs at this level when the listener correctly hears the words but misidentifies which act is being performed through them.

In addition to the three aspects, speakers also weigh face-threatening acts in proportion to social distance, power differences, and the degree of imposition (Brown & Levinson, 1987). In their work, the weight of a face-threatening act is expressed as:

$$W_x = D(S, H) + P(H, S) + R_x$$

where $W_x$ is the weight of the act, $D$, the social distance between speaker and listener, $P$, their power difference, and $R_x$, the culturally defined ranking of imposition. The authors present this additive formula as an approximation. When speakers and listeners assess these factors differently, indirect wording can obscure the intended act.

Another major focus on misunderstanding, the Cross-cultural Pragmatics, extends this problem across speech communities, where the conventions for expressing requests, refusals, and warnings differ. This has been described as pragmatic failure, and two varieties of it have been distinguished (Thomas, 1983). Pragmalinguistic failure is a mismatch in how an act is conventionally expressed, whereas sociopragmatic failure is a mismatch in judgements about what is socially appropriate. Notably, neither variety requires either party to lack competence in the language being used.

This literature contributes a failure that concerns the act performed rather than the content conveyed, and we place it at the same stage as the previous mechanism. The listener correctly hears the words but misidentifies the act performed through them, and we carry that forward as Illocutionary Force Mismatch. The cross-cultural work reviewed here also strengthens the case for Pragmatic Inference Failure, since a convention that softens an act is recovered by inference rather than from the words alone.

## 2.4 Grounding Theory and Common Ground

Common ground is the knowledge, beliefs, and assumptions that two participants share and know that they share (Clark & Brennan, 1991). In their work, communication proceeds when each contribution is understood well enough for the participants' current purposes, rather than completely. Common ground is built incrementally, through uptake, acknowledgement, and clarification (Clark & Schaefer, 1989). However, that same incremental process can certify a divergence rather than resolve it, because apparent acceptance may mask a misunderstanding and silence is readily taken for agreement.

Reference is established collaboratively rather than through a single act of naming (Clark & Wilkes-Gibbs, 1986). In their research, one participant proposes a way of referring and the other accepts, refines, or repairs that proposal. They have shown that misalignment arises when this process terminates before the two participants have converged on the same referent.

Common ground has also been modelled formally, as a context set of propositions that the participants jointly take for granted (Stalnaker, 1973, 2002). When two parties operate with different context sets, the same utterance may carry different presuppositions and implications for each of them. Notably, this divergence produces no surface disturbance, and neither participant need notice that it has occurred.

Recent corpus work has begun to measure this divergence directly (N. Li et al., 2026). In their study, the intended referent and the interpreted referent were annotated separately across roughly 13,000 expressions in the HCRC MapTask corpus (A. H. Anderson et al., 1991). They have shown that apparent grounding can conceal referential divergence, particularly where the two participants' maps show different numbers of the same landmark. However, because the annotation relied on a constrained large-language-model pipeline, we treat these findings as suggestive.

Grounding theory further holds that different communication media supply different grounding resources. Constraints on feedback, visibility, and the opportunity to repair therefore shape which misunderstandings arise and which persist. We return to this point in Section 2.8.

Our analysis separates three processes here rather than one, and places them at two different layers. At the layer where the semantic and pragmatic alignment of an utterance is settled, two parties may attach the same expression to different referents, which we carry forward as Referential Misalignment, and an utterance may take for granted a proposition the listener does not hold, which we carry forward as Presupposition Mismatch. Earlier than either, at the layer where each party estimates what is shared, a participant may proceed as though knowledge were held in common when it is not, which we carry forward as Common-Ground Overestimation. However, the three differ in scope rather than in kind, and we set out the boundaries between them where the mechanisms are defined.

## 2.5 Conversation Analysis and Repair

Repair has been described as an organized system for handling recurring problems in speaking, hearing, and understanding (Schegloff et al., 1977). In their account, four types are distinguished: self-initiated self-repair, self-initiated other-repair, other-initiated self-repair, and other-initiated other-repair. They have shown that the system carries a general preference for speakers correcting themselves.

Repair is pervasive across languages, and it is frequent enough to be measured. In a survey of twelve languages, repair was found to be initiated on average about once every 1.4 minutes, in similar forms throughout (Dingemanse et al., 2015). Repair has also been described as an interface between different approaches to human interaction, because it supplies observable evidence that a communicative problem has occurred (Albert & de Ruiter, 2018).

The same interactional machinery imposes a timing constraint that bears directly on how utterances are produced. In a survey of ten languages, ranging from small indigenous communities through to major world languages, the median gap between turns was found to fall between zero and three hundred milliseconds (Stivers et al., 2009). Notably, that interval is very short in comparison to what production requires, since the planning of even a one-word utterance occupies at least six hundred milliseconds (Holler et al., 2015). In their account, a speaker is consequently already encoding a response while the incoming turn remains unfinished, and must project its ending in order to launch on time. However, the consequence is not confined to timing. A speaker who commits to an utterance before the other turn has closed also commits before the message behind it has been fully specified.

A further distinction separates problematic reference from problematic sequential implicativeness (Schegloff, 1987). The latter covers uncertainty about whether an utterance is serious, how a sequence favours particular responses, whether a turn is heard as a whole or in parts, and how an opening joke frames what follows. Notably, the same work shows that the organization of repair is largely independent of the type of trouble that produced it.

Repair need not wait for trouble to surface, because it can also operate pre-emptively. In their work, speakers expand or clarify a contribution before any misunderstanding becomes visible in the listener's response (Raymond & Gill, 2025).

Repair is also sensitive to who the listener believes they are addressing. In their study, participants held dyadic conversations with a conversational agent, either through a text interface or through a human body relaying agent-generated words, and either knowing or not knowing that the words were agent-generated. They have shown that participants initiated repair less frequently when the exchange ran through a text interface, and less frequently again when they knew the words were agent-generated (Corti & Gillespie, 2016). However, this is a finding about the willingness to repair rather than about the capacity to do so, and the two come apart when one party is not a person.

Two situations have been identified in which unequal access to information obstructs repair (Suchman, 1987). In a false alarm, a system identifies trouble where none exists. In a garden path, an error remains hidden until its source has become difficult to reconstruct. However, both show the same thing, which is that a divergence persists whenever the evidence needed to recognise it is unavailable to the party who would otherwise raise it.

This literature contributes something the others do not. We place it not at any stage that produces a divergence but at the regulation and recovery that follows one, the machinery by which a divergence becomes visible and is resolved. We carry its absence forward as Repair Failure. The same literature also documents the two troubles it most often has to resolve, problematic reference and uncertainty about what an utterance is doing, and so supports Referential Misalignment and Illocutionary Force Mismatch as well.

## 2.6 Cognitive Psychology: Perspective-Taking, Alignment, and Egocentric Bias

Listeners often interpret a message from their own perspective rather than the speaker's (Keysar et al., 2000). In the director-array task, listeners took into account objects that the speaker could not see, which indicates that egocentric interpretation is the default and that perspective-taking requires additional effort. Evidence of the same bias has been found in grammatical perspective-taking, and it is particularly marked in production (C. J. Anderson & Dillon, 2023).

Three related biases contribute to misunderstanding:

- Illusion of transparency: speakers overestimate how visible their thoughts and intentions are to others, and therefore supply less information than listeners actually need (Gilovich et al., 1998).
- Curse of knowledge: once people know something, they struggle to reason from the perspective of someone who does not (Camerer et al., 1989). In a tapping study, participants greatly overestimated how often listeners would identify familiar tunes (Newton, 1990).
- Illusion of explanatory depth: people believe they understand something more fully than they do, until they are asked to explain it, and the result can be an incomplete or poorly specified message (Rozenblit & Keil, 2002).

The interactive alignment model proposes that dialogue depends additionally on automatic convergence in sounds, words, and grammatical structures (Pickering & Garrod, 2004). When that alignment breaks down, misunderstanding can arise even where neither party has failed to consider the other's perspective.

Production research bears on the same egocentric tendency from the speaker's side. In their experiments, eye-tracked speakers incorporated new elements into an utterance after articulation had already commenced, which indicates that a fully specified message is not a precondition of speaking (Brown-Schmidt & Konopka, 2015). The analysis of spontaneous self-repair points in the same direction. In their corpus of nine hundred and fifty-nine repairs, speakers were found to interrupt an utterance in progress in order to replace the message being expressed, and not merely to correct an error in the manner of its expression (Levelt, 1983). Organizational sensemaking research treats meaning and intention in comparable terms, as developing through interpretation rather than being fully formed in advance (Weick, 1995). However, none of this work was conducted in order to explain misunderstanding, and we draw upon it for the condition it establishes rather than for any demonstration of divergence.

We distinguish three processes in this literature, and place two of them at the same layer, where each party forms a view of the other's perspective and of the common ground between them. A participant may interpret or compose from their own viewpoint although the other's is available to them, which we carry forward as Perspective-Taking Failure. The curse of knowledge, the illusion of explanatory depth, and the illusion of transparency all bear instead on the mistaken belief that background is shared, whether the background is factual or is the speaker's own state of mind, and we carry those forward as Common-Ground Overestimation. A third process sits earlier still, at the formation of the intended meaning itself, where a speaker sets out before that meaning has settled. We carry it forward as Intent Underspecification. Its support comes from the production research reviewed above, from the timing of conversational turns set out in Section 2.5, and from organizational research into how meaning is settled through interpretation.

## 2.7 Social Psychology: Attributional Distortion

The fundamental attribution error is the tendency to explain another person's behaviour through their character while underweighting the situation (Heider, 1958; Ross, 1977). In communication, an unclear or incomplete message may therefore be read as carelessness, hostility, or disrespect, rather than as an ordinary communication failure.

The false consensus effect is the tendency to overestimate how widely one's own beliefs and assumptions are shared (Ross et al., 1977). Each party may therefore assume that what is obvious to them is equally obvious to the other, and neither has reason to check.

Attributional judgements often follow an initial misunderstanding and make it harder to repair, because they convert a communication problem into a judgement about the person. However, they can also produce the initial divergence, in cases where the listener misreads the speaker's motive, intent, or relational stance.

We place the process this literature contributes at the stage of relational interpretation, where a communicative difficulty is read as evidence about the person rather than about the message. We carry it forward as Attributional Distortion. The false consensus effect reviewed here bears additionally on Common-Ground Overestimation, since projecting one's own beliefs onto another is one route to believing they are shared.

## 2.8 Medium and Channel Effects

Media richness theory holds that media differ in their capacity to reduce misinterpretation, through rapid feedback, personalisation, multiple cues, and natural language (Daft & Lengel, 1984, 1986). Lean media therefore raise the risk of misunderstanding when a message is ambiguous or emotionally charged. Nonverbal cues have further been shown to carry meaning that depends on context, rather than being readable as a fixed code (Patterson et al., 2023). They also carry information that the words themselves do not. Listeners who could see a speaker's iconic gestures recovered the relative position and size of objects more accurately than listeners who heard the same speech alone (Beattie & Shovelton, 1999). Prosody performs comparable work, and speakers produce prosodic boundaries that allow a listener to resolve a syntactic ambiguity before the ambiguous phrase is complete (Snedeker & Trueswell, 2003). However, the gesture effect reached significance only for particular semantic categories, so the loss of a nonverbal channel is better understood as removing some information than as removing a fixed proportion of the message.

Miscommunication during collaborative tasks has been found to be associated with more ambiguous utterances and lower lexical density (Paxton et al., 2021). In their study, the more successful pairs used denser wording at precisely the points where they were grounding each other's contributions.

Technology can remove the opportunities for mutual monitoring and repair that people rely on without noticing (Heath & Luff, 2000). In their account of the London Ambulance Service computer-aided dispatch system, calls were lost, delayed, or duplicated once the system replaced parts of the earlier radio-and-telephone coordination process.

The medium usually increases the effect of another communication failure. However, it can also create a divergence directly, when it alters what the listener receives, as with dropped audio, garbled transcription, a missing attachment, a delayed message, or a cropped image.

This literature contributes a process that is usually secondary and occasionally primary. Our analysis places it at the medium itself, and at the interaction constraints the medium imposes. A medium may withhold, degrade, delay, or alter what reaches the listener, and we carry that forward as Channel and Medium Distortion. The corpus evidence reviewed here, linking miscommunication to ambiguous wording and low lexical density, also bears on Encoding Ambiguity.

## 2.9 Intercultural Communication

The high-context and low-context distinction concerns how much meaning is expressed directly and how much is inferred from shared background (Hall, 1976). However, because it can treat cultures as uniform and context as a simple binary, we use it only as an orienting concept.

Five dimensions have been identified along which communicative norms vary (House, 2006). These are direct versus indirect, self-oriented versus other-oriented, content-focused versus addressee-focused, explicit versus implicit, and improvised versus routine.

The pragmalinguistic and sociopragmatic distinction introduced in Section 2.3 applies directly here (Thomas, 1983). Such failures have been linked to differences in values, cultural practices, patterns of thought, and transfer from a speaker's first language (Q. Li & Cao, 2019). For example, routine Chinese greetings like "Have you eaten?" may be read by English speakers as intrusive rather than as ordinary conversation openers.

A further variety, malaprop-pragmatic failure, arises where an otherwise appropriate utterance misfires because of a slip under pressure (Lu, 2019). Additional sources have been identified in written communication, including bluntness, missing softening language, omitted greetings and closings, and inappropriate forms of address (McGee, 2019).

Intercultural difference is not itself a single mechanism. It is a condition that raises the likelihood of mismatched assumptions, inferences, speech acts, and expectations, while at the same time discouraging repair, because admitting a misunderstanding can be face-threatening.

Unlike the other eight, we find no mechanism of its own in this literature. What it describes is a condition under which several of the mechanisms above become more likely, and under which repair becomes harder to initiate. However, the cross-cultural pragmatics reviewed here does supply further evidence for two of them, Pragmatic Inference Failure and Illocutionary Force Mismatch, since the conventions that vary across speech communities are precisely those governing what is implied and which act is performed.

Across the nine fields, misunderstanding is examined through different units of analysis, including signal transmission, pragmatic inference, speech acts, grounding, repair, cognitive bias, attribution, medium constraints, and cultural conventions. However, despite these differences, the fields converge on a common structure. That convergence is what makes a single synthesis possible.

## 2.10 Prior Classifications of Misunderstanding

In this section, we turn from the fields that supply the mechanisms to the classifications of misunderstanding that already exist. Misunderstanding has also been classified in its own right, however, and this synthesis is not the first taxonomy of it. Setting out what already exists is what makes the remaining work visible.

One of the earliest systematic classifications identified in this review separates misperception from misinterpretation, then locates each at the phonological, syntactic, semantic, or situational level, and asks separately whether the illocutionary force or the propositional content is affected (Zaefferer, 1977). A pragmatics-based account reduces the question to two levels, the proposition and the pragmatic force, and it is failure at the second that gives us pragmatic failure (Thomas, 1983). A dialogical account classifies by visibility instead (Linell, 1995). An overt misunderstanding is recognised at once, a covert one only later, and a latent one is never recognised by the participants at all.

Later work varies the dimensions rather than the object. One taxonomy is built from three continua, intentional against unintentional, verbal against nonverbal, and explicit against implicit (Yus Ramos, 1999). Another turns from classifying misunderstandings to the interactional work of handling one once it surfaces (Bazzanella & Damiano, 1999). A multidimensional model is organized around a message transfer circle and the mental worlds of the two participants (Mustajoki, 2012). That account argues that ambiguity and misreference are only risks of miscommunication. The cause it identifies is incomplete recipient design. A multidisciplinary review sets out causes, factors, and consequences across disciplines without reducing them to a single scheme (Padilla Cruz, 2023).

A second body of work classifies not the misunderstanding but the level at which joint action fails. That structure is set out as a ladder of four levels (Clark, 1996). The speaker executes behaviour, presents a signal, signals that something is the case, and proposes a joint project, while the addressee attends, identifies, recognises, and considers. Two properties hold the ladder together. Upward completion means that completing any level requires completing every level beneath it. Downward evidence means that evidence of success at one level is evidence of success at all levels beneath it. Grounding, on this account, is required at every level rather than at the end. The eight layers of Section 4 are a descendant of this idea, at finer resolution and derived from a wider evidence base.

A third classifies errors in machine dialogue. Seventeen error types have been proposed across four levels, being the utterance, the response, the context, and society, constructed by integrating a theory-driven taxonomy with a data-driven one (Higashinaka et al., 2021). That integration is the strategy this synthesis also uses, applied there within one field rather than across nine. Failures in human-robot interaction have been divided into technical and interaction failures, alongside a model that stages the handling of a failure as human information processing, running from how the failure is communicated, through how it is perceived and understood, to how it is solved (Honig & Oron-Gilad, 2018). A corresponding taxonomy exists for multi-agent systems built on large language models (Cemri et al., 2025).

Against this background the contribution can be stated precisely. This is not the first taxonomy of misunderstanding, not the first model to locate a failure at a level, and not the first to integrate taxonomies built by different methods. Three things are new. The first is breadth. Each of the classifications above stays within a single field, whether pragmatics, joint action, or machine dialogue. The mechanisms here are drawn from nine fields, and they cover communication between people and between people and machines under one scheme. The second is that we separate where a failure occurs from what it does. A mechanism is placed at a layer and independently typed by function, as generating a divergence, amplifying one already present, or governing whether one is detected. We have found no prior taxonomy that makes both assignments. The third is that the mechanisms are given a formal statement in Section 4.2 rather than remaining, however comprehensive, a list of named categories.

The debts are as clear as the differences, and we record them. The overt, covert, and latent misunderstandings of the dialogical account prefigure the detection and resolution states we set out in Section 3 (Linell, 1995). The ladder of joint action prefigures the layers (Clark, 1996). Recipient design prefigures the layer at which each participant estimates what the other knows (Mustajoki, 2012), and the convergence goes further than the name. That account treats ambiguity and misreference as risks rather than causes, holding that the cause lies upstream in an incomplete model of the recipient, and our second layer behaves the same way, since the estimate formed there conditions formulation, alignment, inference, and relational interpretation alike. We stop short of the hierarchy it proposes. Our structure stays flat because a mechanism at a later layer can produce a material divergence with the conditioning state perfectly correct, and treating every such case as downstream of recipient design would obscure that. What none of this earlier work supplies, however, is a definition precise enough to decide whether a given divergence counts as a misunderstanding at all, and that is where Section 3 begins.

# 3. Defining Misunderstanding

## 3.1 The Definition

The fields reviewed in Section 2 differ in terminology and analytical focus, but they converge on a common structure. In every case, one participant reconstructs a meaning that is materially different from the meaning another sought to communicate. The divergence may concern a referent, an implied meaning, a communicative act, a relational stance, or a requested action. On that basis, we adopt the following working definition.

*A misunderstanding is a divergence between the meaning a speaker sought to communicate and the meaning a listener reconstructed. The divergence may concern the intended referent, an implied meaning, the speech act being performed, the relational stance being conveyed, or the action being requested.*

This definition was developed for the purposes of the synthesis. It does not replace the more specialized definitions used within the individual fields we reviewed.

Before stating the criterion formally, we set out the chain of transformations that a communicative episode passes through, shown in Figure 1. The speaker's intended meaning is encoded into a signal, and we write that encoding as E. It is driven by the speaker's world model and by the goal being pursued, rather than by a fixed code. The signal itself we write as X. The channel, which we write as C, transforms X into the signal the listener actually receives, shown in Figure 1 as X with a hat. The listener then reconstructs a meaning from that received signal, and we write that reconstruction as D, driven in turn by the listener's own world model and by the action they are preparing to take. D is not a fixed decoder. It is an inference process, and Section 4.2 opens it into a posterior over candidate meanings. Figure 1 also shows a return path, drawn as a dashed line. Repair is not a separate channel running alongside the forward one. Interactive repair traverses the same chain a second time with the roles reversed, so that a correction is encoded, transmitted, and reconstructed exactly as the original message was. Self-initiated repair and pre-emptive clarification differ, because they intervene before the forward traversal is complete rather than reversing it. In each case the repair travels the same path as the message and is therefore exposed to the same mechanisms. We draw it dashed because its availability is not guaranteed, and we treat its absence in Section 4 as a mechanism in its own right.

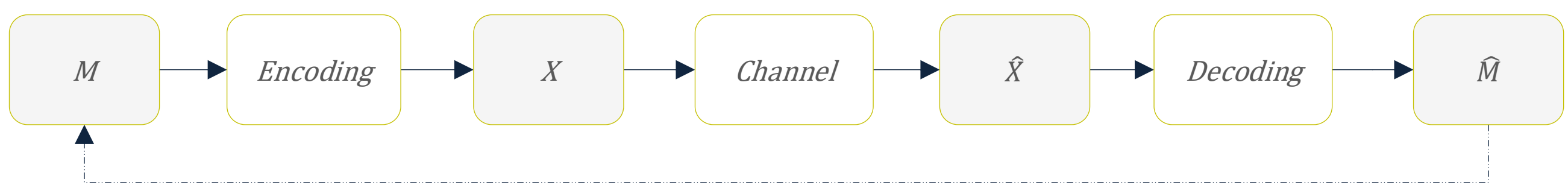

*Figure 1. The communicative chain. A meaning is encoded into a signal, transformed by the channel, and decoded into a reconstructed meaning. The dashed return path is repair, shown schematically. Interactive repair runs as a second traversal of the same chain with the roles reversed, whereas self-initiated and pre-emptive repair intervene before the forward traversal completes.*

The channel imposes three kinds of transformation, and they are worth separating because they fail in different ways. Signal power is attenuated, some content is removed selectively rather than uniformly, and noise is added. In air the second of these is frequency-dependent, so the channel behaves as a low-pass filter (*ISO 9613-1*, n.d.), and it is therefore the transformation that removes distinguishable content rather than merely level. A listener can compensate for attenuation by moving closer or by asking for more volume. However, once the affected content falls below the effective noise floor, broadband gain does not restore it, because gain raises the signal and the noise together. Section 4.2 treats the medium as Layer 4 and gives the channel its formal statement there.

The acoustic case is one instance of the general form, in which the received signal is a transformation of the transmitted one together with added noise. Where an exchange is written, C is the transformation imposed by the medium and its handling. Geometric spreading and molecular absorption have no counterpart there, but attenuation, selective loss, and added noise do, appearing as truncation, transcription error, formatting loss, and delay.

The departure from Shannon lies in E and D. In Shannon's formulation, the encoding and decoding rules are known and fixed within the system, and the coding theorem establishes that under those assumptions transmission can be made arbitrarily reliable at any rate below channel capacity. In an exchange between two people, E and D are private. Each is a function of a world model the other party cannot inspect, and the two world models are free to diverge without either party becoming aware of it. Channel capacity therefore does not by itself constrain the divergence between the intended and the reconstructed meaning, because that divergence is introduced by mismatched encoding and reconstruction functions that lie outside the technical channel model. This is the reason a mechanism-level account is required, and the reason a single noise term will not serve in place of one.

Shannon's distinction between transmitted and reconstructed signals can be extended conceptually from transmitted signals to reconstructed meanings. Let $M$ denote the meaning the speaker sought to communicate and $\widehat{M}$ the meaning reconstructed by the listener. A misunderstanding occurs when the two meanings differ materially, where $d$ is a semantic-distance or task-loss function:

$$d(M, \widehat{M}) \geq \tau,$$

The threshold $\tau$ is the point at which the difference becomes material. When meanings are represented as discrete categories, this criterion reduces to:

$$M \neq \widehat{M}.$$

Across a collection of episodes, the prevalence of misunderstanding may be expressed as:

$$P_{\text{mis}} = \Pr(M \neq \widehat{M}),$$

while its average magnitude may be represented as:

$$D_{\text{mis}} = \mathbb{E}[d(M, \widehat{M})].$$

A Shannon-style conditional entropy,

$$H(M \mid \widehat{M}),$$

can describe the residual uncertainty about intended meaning given reconstructed meaning across a population of episodes. It does not, by itself, determine whether a particular episode constitutes a misunderstanding. The episode-level criterion remains a material divergence between $M$ and $\widehat{M}$.

The threshold is defined by consequence and not by detection. A divergence crosses it when it would lead the listener to take a different action, form a different commitment, or draw a different conclusion from the one the speaker sought. Defining it in this way keeps detection out of the definition, and it preserves the separation set out below, in which Repair Failure governs whether a divergence becomes visible rather than determining whether it counts as a misunderstanding at all.

Materiality is not the only quantity of interest, and it is not the one that determines whether a misunderstanding proves survivable. A second quantity describes how readily an existing divergence can be surfaced and corrected. We refer to it as recoverability. It is distinct from the magnitude of the divergence and does not follow from it. Recoverability depends on whether the interaction produces evidence of divergence, whether that evidence is observable to either participant, and whether sufficient feedback and repair actions remain available. A larger semantic distortion need not imply lower recoverability, because a large divergence may produce stronger observable inconsistencies than a small one, whereas a small but consequential divergence can leave the interaction looking entirely ordinary. What does reduce recoverability is the withdrawal of feedback, visibility, timing, and the opportunity for immediate clarification, because those are the resources through which repair operates. We formalise recoverability through observation and updating in Section 4.2.

Recoverability is deliberately held outside the definition. A divergence does not cease to be a misunderstanding because it is easy to repair, and it does not become one because it is hard. However, separating the two quantities gives the three functional roles of Section 4.3 a formal shape. A generating mechanism creates a divergence where there was none. An amplifying mechanism increases that divergence, degrades the resources on which recoverability depends, or does both. A detecting mechanism governs recoverability directly. We do not instantiate the distance function, the threshold, or the recoverability function here, and we return to that obligation in Section 6.5.

Detection and resolution are analytically separate from the existence of the divergence. Conversation analysis distinguishes a trouble source from the repair practices through which it becomes visible and may be resolved (Albert & de Ruiter, 2018; Schegloff, 1987). A misunderstanding may therefore remain undetected, be detected but unresolved, or be detected and resolved. Including detection in the definition would create a circularity, in which a divergence ceased to qualify as a misunderstanding once it was recognised. It would also make Repair Failure a defining condition, rather than a mechanism governing whether an existing divergence becomes visible and is addressed.

References to intended meaning do not imply direct access to a speaker's private mental state. Intended meaning must be supported by observable evidence, e.g. an explicit correction, a clarification, a subsequent action, an independently established referent, task structure, or a participant account (Heritage, 1984; Schegloff, 2007). In constructed cases, we specify the intended meaning as part of the illustration. The taxonomy therefore analyses evidence of divergent understandings, rather than assigning intentions on the basis of the analyst's interpretation alone.

## 3.2 Distinguishing Misunderstanding from Related Concepts

The working definition distinguishes misunderstanding from several related phenomena. In particular, ambiguity alone is a property of a message. It becomes Encoding Ambiguity only when the listener selects a meaning different from the one intended.

*Table 1. Conceptual Boundaries of Misunderstanding in This Synthesis*

| CONCEPT | DISTINGUISHING FEATURE | TREATMENT IN THIS SYNTHESIS |
|---|---|---|
| **MISHEARING** | A sound or perceptual failure that occurs before meaning is interpreted | Included only when it produces a divergence in the meaning reconstructed by the listener |
| **AMBIGUITY** | A message permits more than one plausible interpretation | Excluded by itself, because it is a property of the message rather than an outcome |
| **ENCODING AMBIGUITY** | The mechanism is present when the listener actually selects a reading different from the one the speaker intended | Included as a mechanism that generates misunderstanding |
| **DISAGREEMENT** | The message is understood correctly, then rejected | Excluded |
| **DECEPTION** | The speaker deliberately seeks to induce a false belief | Excluded, unless the listener also misreads the deceptive intent itself |
| **MISUNDERSTANDING** | The listener reconstructs a meaning different from the one the speaker sought to communicate | The core phenomenon, subsequently classified by its detection and resolution state |

## 3.3 Detection and Resolution States

Every misunderstanding can be classified by two sequential questions, being whether the divergence was detected, and if it was, whether it was resolved. These questions produce three possible states.

1. **Undetected: Neither participant recognises the divergence.**
2. **Detected but unresolved:** The divergence becomes visible, but repair fails or remains incomplete.
3. **Detected and resolved:** The divergence becomes visible and repair restores sufficient mutual understanding.

Detection and resolution classify what happens after a divergence has occurred. They do not determine whether the divergence qualifies as a misunderstanding. This broader usage includes both repaired and unrepaired misunderstandings, and it avoids making Repair Failure part of the definition, which would create the circularity discussed in Section 3.1.

Undetected misunderstandings may be especially consequential, because they can influence subsequent actions without either participant recognising the problem. They are also the hardest to observe. Evidence is therefore strongest for misunderstandings revealed through correction or repair, and weakest for those that remain unnoticed. However, this is a limitation of the available evidence rather than of the phenomenon, and we return to it in Section 6.6.

This classification is consistent with the conversation-analytic view that repair restores sufficient intersubjectivity and allows the interaction to progress, rather than necessarily making the participants' mental representations identical (Albert & de Ruiter, 2018).

# 4. The Taxonomy: Cross-Disciplinary Conceptual Synthesis

Using best-fit framework synthesis (Carroll et al., 2013), we compared recurring concepts across the nine fields reviewed. We combined overlapping accounts, and we retained separate mechanisms where they differed in where or how they operated. This section presents the result in three parts, and the order of presentation is the order of derivation. We begin with the eleven mechanisms, taken at the eight points in a communicative episode where our analysis places them. We then state those eight points formally, as the chain a meaning passes through on its way from one participant to another. Only with both in view do we turn to the three functional roles, which are what the completed picture shows rather than a scheme brought to it.

## 4.1 The Eleven Mechanisms and Where They Operate

The literature reviewed in Section 2 proposes lots of failure mechanisms under lots of names, and the density of the naming is itself part of the problem. Consolidating those that describe the same process left eleven distinct mechanisms. Analysing each in turn showed that it does not operate anywhere in a communicative episode but at a particular point within it, and that the eleven fall between only eight such points. We call those points layers, and as can be seen from Table 2 in Section 4.2, all eight are listed together there. Here we take them in order, defining each and introducing the mechanisms that operate there. Five layers admit a single mechanism and three admit two, and we return to that asymmetry at the end of the section.

The first layer is Formation, where a communicative goal becomes a determinate intended meaning. One mechanism operates here.

### 1. Intent Underspecification

Intent Underspecification generates a divergence when the speaker has not formed a sufficiently determinate meaning before attempting to express it. The intended meaning may be vague, unstable, internally inconsistent, or incomplete.

The evidence reviewed in Sections 2.5 and 2.6 supplies the basis for this mechanism. Speaking does not wait upon a fully specified message, and the timing of conversational turns renders early commitment the ordinary case rather than the exception. Cognitive overload, unfamiliar or complex tasks, and time pressure compound a constraint that the structure of conversation already imposes. In an episode the mechanism must usually be inferred from later behaviour, such as substantial revision, contradictory explanations, or an inability to clarify what was originally meant.

The second layer is Perspective and Common Ground, where each participant estimates what the other knows and can be expected to recover, and puts that estimate to use. Two mechanisms operate here. However, this layer does not act only at one moment. The estimate formed here continues to shape how the message is worded, how its referents are settled, how it is interpreted, and how it is read for relational meaning.

### 2. Common-Ground Overestimation

Common-Ground Overestimation generates a divergence when one or both participants proceed as though particular knowledge, beliefs, assumptions, or background information were shared when they are not.

The mechanism reflects a failure to establish or update common ground through grounding (Clark & Brennan, 1991). It is reinforced by three well-documented biases. The false consensus effect leads people to project their own beliefs onto others, the curse of knowledge makes it difficult to reason from the position of someone who lacks what one knows, the illusion of explanatory depth produces unwarranted confidence in the clarity of one's own understanding (Camerer et al., 1989; Ross et al., 1977; Rozenblit & Keil, 2002), and the illusion of transparency leads speakers to overestimate how evident their own intentions are, so that they supply less than the listener needs (Gilovich et al., 1998).

### 3. Perspective-Taking Failure

Perspective-Taking Failure generates a divergence when a participant interprets or formulates a message from their own viewpoint rather than the other person's, even when information about the other perspective is available.

Egocentric interpretation is often the fast default, while incorporating another person's perspective requires additional processing and may fail under pressure (Keysar et al., 2000). Dialogue also depends on automatic alignment in words and grammatical structures. When that alignment breaks down, misunderstanding can arise without a deliberate failure to consider the other person's position (Pickering & Garrod, 2004). The asymmetry noted in Section 2.6, that the bias falls more heavily on production than on comprehension, matters here, because it places the mechanism on the speaker's side as often as on the listener's (C. J. Anderson & Dillon, 2023).

The third layer is Formulation, where the intended meaning is expressed in words or other signals, under the estimate formed at the previous layer. One mechanism operates here.

### 4. Encoding Ambiguity

Encoding Ambiguity generates a divergence when the speaker expresses an intended meaning in a form that permits more than one plausible interpretation and the listener selects a different interpretation from the one intended.

It may arise through words with multiple senses, unclear syntax or scope, uncertain pronoun reference, unstated thresholds in terms such as "soon," or jargon without a shared definition. The corpus evidence reviewed in Section 2.8 supplies the empirical anchor, since the wording that accompanied miscommunication there was measurably less dense at the points where grounding was being done (Paxton et al., 2021). Communication theory also identifies encoding fidelity as a distinct point of possible failure (Shannon, 1948). Computational accounts also treat ambiguity and clarification as recurring forms of miscommunication in dialogue (Purver et al., 2018).

The fourth layer is Medium and Interaction Constraints, which governs the transmission of cues and the grounding resources available to the participants. One mechanism operates here.

### 5. Channel and Medium Distortion

Channel and Medium Distortion characteristically worsens a divergence already present, and it does so when the communication medium removes, degrades, delays, or alters information needed to interpret and ground a message. Relevant resources include timing, tone, facial expression, gesture, visibility, shared physical context, and opportunities for immediate clarification.

Section 2.8 reviews the evidence. Lean media provide fewer resources for resolving ambiguity (Daft & Lengel, 1984, 1986), and different media withdraw different grounding resources (Clark & Brennan, 1991), while nonverbal meaning depends on context rather than being readable as a fixed code (Patterson et al., 2023). Losing a nonverbal channel therefore removes information the words do not supply, as with gestural cues that improve recovery of relative position and size (Beattie & Shovelton, 1999), and prosodic boundaries that resolve a syntactic ambiguity in speech but are absent from text (Snedeker & Trueswell, 2003).

The fifth layer is Semantic and Pragmatic Alignment, where referents are settled and the propositions an utterance takes for granted are resolved. Two mechanisms operate here.

### 6. Referential Misalignment

Referential Misalignment generates a divergence when the speaker and listener connect the same expression to different objects, people, events, times, locations, or states.

Reference is established collaboratively. One participant proposes a referent, and the other accepts, refines, or repairs it (Clark & Wilkes-Gibbs, 1986). Misalignment arises when apparent acceptance occurs before both participants have identified the same referent.

The mechanism is supported by experimental work on collaborative reference, conversation-analytic evidence of problematic reference, and corpus research indicating that apparent grounding can conceal different referent selections (Clark & Wilkes-Gibbs, 1986; N. Li et al., 2026; Schegloff, 1987).

### 7. Presupposition Mismatch

Presupposition Mismatch generates a divergence when an utterance assumes a proposition that the listener does not share. The listener may accept the assumption without challenge or may be unable to form a complete interpretation.

Presuppositions can be introduced through factive verbs, definite descriptions, cleft constructions, particles, and change-of-state verbs. The listener may accommodate the presupposition by adding it to the conversational context even when it was not previously shared (Lewis, 1979; Stalnaker, 1973). Discourse-semantic accounts further model presupposition resolution as an anaphoric process within discourse context (Van Der Sandt, 1992).

Experimental evidence indicates that accommodating an unestablished presupposition carries greater processing cost than interpreting one already supported by the context (Chemla & Bott, 2013).

The sixth layer is Inference and Interpretation, where implied meaning and the communicative act being performed are reconstructed. Two mechanisms operate here.

### 8. Pragmatic Inference Failure

Pragmatic Inference Failure generates a divergence when the listener reconstructs an implied or indirect meaning different from the one the speaker intended.

It may arise because the participants rely on different background assumptions when applying conversational expectations (Grice, 1975), evaluate relevance against different contexts (Sperber & Wilson, 1995), or assign different salience to competing meanings (Giora, 1997, 2003).

### 9. Illocutionary Force Mismatch

Illocutionary Force Mismatch generates a divergence when the listener misidentifies the communicative act being performed. A request may be interpreted as a question, a warning as a threat, a joke as a sincere assertion, or a refusal as an invitation to continue negotiating.

The mechanism is grounded in the distinction between the words uttered and the act performed through them (Austin, 1962; Searle, 1969). It becomes more likely when acts are expressed indirectly, softened to manage face, shaped by different cultural conventions, or interpreted through misleading sequential expectations (Brown & Levinson, 1987; House, 2006; Schegloff, 1987; Thomas, 1983).

The seventh layer is Relational Interpretation, where motive, disposition, and relational stance are read from the same material. One mechanism operates here.

**10. Attributional Distortion**

Attributional Distortion characteristically worsens a divergence already present, and it does so when a participant explains another person's communicative behaviour through character, motive, or disposition while underweighting the situation or the ordinary difficulty of communicating.

The mechanism is grounded in research on dispositional attribution and the fundamental attribution error (Heider, 1958; Ross, 1977). An unclear or incomplete message may therefore be interpreted as carelessness, hostility, manipulation, or disrespect.

The eighth layer is Regulation and Recovery, where a divergence is detected, clarified, corrected, and repaired, or is not. One mechanism operates here, and it governs whether anything at the preceding seven layers ever becomes visible.

**11. Repair Failure**

Repair Failure produces no divergence of its own. It governs whether one becomes visible, and it operates when the interaction lacks the checking, clarification, confirmation, correction, or acknowledgement that would expose and address a divergence.

Repair may fail because the divergence never appears on the surface of the interaction, face concerns discourage clarification, the medium restricts feedback, or a repair restores conversational progress without resolving the underlying difference in meaning. Whether it fails is what determines whether a misunderstanding remains undetected, becomes detected but unresolved, or is detected and resolved. Repair may also fail because of who the listener believes they are addressing, since the willingness to initiate repair falls when the interlocutor is known to be an artificial agent (Corti & Gillespie, 2016).

Five of the eight layers admit a single failure mechanism and three admit two. That asymmetry is not an artifact of where we drew the boundaries. The three layers carrying two are the points at which the reviewed literatures themselves maintain a distinction that does not survive being collapsed: between settling which entity an expression picks out and accommodating a proposition an utterance takes for granted, between recovering what was implied and identifying which act was performed, and between holding a mistaken model of the other participant and failing to apply a correct one. Where two candidate mechanisms could be merged without losing a distinction the source literatures draw, the layer carries one. The three two-mechanism layers are therefore where the taxonomic work of this synthesis is concentrated, and they are also where its distinctions most need independent testing.

The eleven definitions above are stated in ordinary language. Each also admits a formal statement in the notation set out in Section 4.2, where a mechanism appears as a perturbation of the nominal conditional distribution at one or more layers. However, this paper develops the mechanism-level account rather than its full formal expression, and we develop that expression separately.

Mechanism identifiers run in process order, so a lower number marks an earlier point in a communicative episode. They are nonetheless flat labels rather than positions within a layer. A mechanism keeps its identifier, however, if later work reassigns it to a different layer, or if a layer is added, so that a revision to the structure does not invalidate every reference to a mechanism.

## 4.2 The Layers as a Formal Probabilistic Communication Model

Section 4.1 identified the eight analytical layers and placed each mechanism at the layer where it principally operates. This section gives those layers a common formal representation. The objective is not to redefine them as compulsory chronological stages. It is to show how the functions identified by the taxonomy can be embedded in an end-to-end probabilistic model of communication in which intended meaning is latent, signals are observed, interpretations are inferred, and repair supplies additional information when divergence becomes visible.

Table 2 summarizes the eight layers.

*Table 2. The Eight Analytical Layers and Their Principal Locations in the Communication Process*

| LAYER | NAME | PRINCIPAL LOCATION |
|---|---|---|
| 1 | Formation | Development of the intended meaning or communicative goal |
| 2 | Perspective and Common Ground | Estimation and use of the other participant's knowledge and viewpoint |
| 3 | Formulation | Expression of the intended meaning in words or other signals |
| 4 | Medium and Interaction Constraints | Transmission of cues and availability of grounding resources |
| 5 | Semantic and Pragmatic Alignment | Alignment of referents and presupposed propositions |
| 6 | Inference and Interpretation | Reconstruction of implied meaning and communicative acts |
| 7 | Relational Interpretation | Interpretation of motive, disposition, and relational stance |
| 8 | Regulation and Recovery | Detection, clarification, correction, and repair |

The layers were not adopted from an existing transmission model. They were derived from the locations at which the mechanisms identified in the literature principally operate. The formal model therefore follows the taxonomy rather than determining it.

### Starting point: the classical communication architecture

Let $M$ denote the meaning the sender seeks to communicate, $X$ the emitted signal, $Y$ the signal available to the receiver after transmission, and $\widehat{M}$ the meaning reconstructed by the receiver. The compact communication architecture is

$$M \xrightarrow{\mathcal{E}} X \xrightarrow{\mathcal{C}} Y \xrightarrow{\mathcal{D}} \widehat{M}.$$

Here, $\mathcal{E}$ is the composite encoding process, $\mathcal{C}$ is the communication channel, and $\mathcal{D}$ is the composite reconstruction process. This retains the structural distinction introduced by information and communication theory between what is produced, what is transmitted, what is received, and what is reconstructed. The extension required for human communication lies primarily in $\mathcal{E}$ and $\mathcal{D}$: neither is a fixed shared codebook, and both depend on private world models, context, expectations, and uncertain beliefs about the other participant.

Accordingly, the receiver is better modelled as inferring a latent intended meaning than as applying a deterministic inverse function to the observed signal.

### Meaning reconstruction as Bayesian inference

Let $\mathcal{M}$ denote the space of candidate meanings. Let $K_L$ denote the listener-side state relevant to interpretation, including available knowledge, assumptions, perspective information, and the current conversational context. After observing $Y$, the listener forms a posterior distribution over candidate meanings:

$$p_L(m \mid Y, K_L, c) \propto p_L(Y \mid m, K_L, c)\, p_L(m \mid K_L, c),$$

where $c$ denotes the remaining communication context.

The first term is the listener's likelihood model: how compatible the observed signal is with candidate meaning $m$ under the listener's model of the situation. The second is the listener's prior over meanings before the current observation is incorporated. Different participants may use different priors, different likelihood models, or different side information even when the physical signal is received perfectly.

Uncertainty in this posterior is not itself misunderstanding. A listener may hold a diffuse posterior and still act on the meaning that was intended, and may hold a sharply concentrated posterior on the wrong one. What the definition in Section 3.1 tests is the distance between the meaning acted upon and the meaning intended, not the confidence with which it was selected.

A maximum-a-posteriori reconstruction is the special case

$$\hat{M}_{\mathrm{MAP}} = \underset{m \in \mathcal{M}}{\operatorname{argmax}} p_L(m \mid Y, K_L, c).$$

However, misunderstanding is consequential because different reconstructions do not carry equal cost. A more general decision-theoretic reconstruction therefore uses the semantic or task-loss function introduced in Section 3.1. Let $\ell_c(M, a)$ denote the loss incurred if the intended meaning is $M$ but the listener acts on reconstruction $a$. The listener's Bayes decision is

$$\hat{M}^{-} = \underset{a \in \mathcal{M}}{\operatorname{argmin}} \mathbb{E}_L[\ell_c(M, a) \mid Y, K_L, c].$$

The superscript minus denotes the reconstruction before any regulation or repair at Layer 8.

This formulation places the paper's materiality criterion inside the established framework of statistical decision theory, in which an action is chosen to minimise expected loss rather than to maximise posterior probability (Wald, 1949). The form used here takes the expectation under the listener's own posterior, which is the Bayesian case (Berger, 1985). The listener may select a reconstruction that is statistically reasonable under their own model and still differ materially from what the speaker sought to communicate because the two participants do not share the same model, prior, context, or side information.

The threshold itself can be read in the same framework. Let the task-relevant consequence of a meaning be the interpretation or action it licenses in the current context. The materiality threshold is then the smallest semantic distortion sufficient to change that consequence, being the infimum of the distortion taken over those meanings whose consequence differs from the one intended. Stated this way the threshold is a decision boundary rather than a free parameter, and it makes formal what Section 3.1 states in words, that a divergence is material when it would lead the listener to take a different action, form a different commitment, or draw a different conclusion.

**Layer 1: Formation as a source distribution**

Let $G$ denote the communicative goal and $W_S$ the sender's world model. Formation produces an intended meaning according to

$$M \sim p_1(m \mid G, W_S, c).$$

When the communicative intention is well settled, this distribution is concentrated. Intent Underspecification corresponds to a source state in which uncertainty remains material at the point formulation begins. One possible measure is

$$H_S(M \mid G, W_S, c),$$

although the taxonomy does not require that every instance of M1 be operationalized by entropy. The important distinction is that Layer 1 concerns uncertainty or instability in the intended meaning before it is encoded into a signal.

**Layer 2: Perspective and Common Ground as side information**

Layer 2 does not behave like an ordinary serial channel. It produces participant-specific conditioning states that alter later probabilities.

Let

$$K_S \sim p_2^S\left(k \mid W_S, \hat{W}_{S \to L}, c\right)$$

represent the sender's working model of what the listener knows, believes, perceives, and can recover, and let

$$K_L \sim p_2^L\left(k \mid W_L, \hat{W}_{L \to S}, c\right)$$

represent the listener's corresponding interpretive state.

The two need not be equal:

$$K_S \neq K_L.$$

This mismatch is not itself a misunderstanding. It is a latent difference in side information that may alter formulation, alignment, inference, or relational interpretation. Common-Ground Overestimation concerns an inaccurate $K_S$, while Perspective-Taking Failure concerns failure to make appropriate use of an available representation of the other participant.

**Layer 3: Formulation as a generative encoding model**

The sender produces an observable expression conditional on intended meaning and the sender-side conditioning state:

$$X \sim p_3(x \mid M, K_S, c).$$

This replaces a fixed encoder with a stochastic formulation model. The same intended meaning can be expressed in several ways, and the same expression may be compatible with several intended meanings.

Encoding Ambiguity is naturally represented by overlapping likelihood support. For two distinct meanings $m_1$ and $m_2$, an expression $x$ is ambiguous when

$$p_3(x \mid m_1, K_S, c) > 0 \quad \text{and} \quad p_3(x \mid m_2, K_S, c) > 0,$$

with the listener subsequently selecting a materially different reconstruction.

**Layer 4: Medium and Interaction Constraints as the communication channel**

The medium maps the emitted signal $X$ to the observation $Y$ available to the receiver:

$$Y \sim p_4(y \mid X, c_4).$$

This is the layer at which conventional signal-processing models remain directly applicable. In an acoustic linear time-invariant special case,

$$Y(t) = h(t) * X(t) + N(t),$$

Here h(t) is the channel impulse response, the star operator denotes convolution, and N(t) is additive noise. Written, visual, or AI-mediated channels require different channel models, but the probabilistic form $p_4(y \mid x, c_4)$ remains general.

Channel and Medium Distortion changes this distribution directly or reduces the grounding information made available to later layers.

**Layer 5: Semantic and Pragmatic Alignment as latent-state estimation**

After receiving $Y$, the listener resolves referents and presupposed propositions against the available context. Let $A$ denote the resulting alignment state:

$$A \sim p_5(a \mid Y, K_L, c).$$

For a referential subproblem with candidate referents $r \in \mathcal{R}$, the listener may select

$$\hat{r} = \operatorname*{argmax}_{r \in \mathcal{R}} p_L(r \mid Y, K_L, c).$$

Referential Misalignment occurs when the selected referent differs materially from the intended one. Presupposition Mismatch changes the context against which the expression is interpreted, and can therefore alter the posterior over admissible meanings even when the received signal is unchanged.

**Layer 6: Inference and Interpretation as probabilistic pattern recognition**

Let $I$ denote the inferred pragmatic interpretation, including implied meaning and communicative force:

$$I \sim p_6(i \mid A, K_L, c).$$

Some components of this layer are naturally categorical. If $F$ denotes illocutionary force, then

$$\hat{F} = \underset{f \in \mathcal{F}}{\operatorname{argmax}} p_L(f \mid A, K_L, c).$$

Similarly, an implied meaning can be treated as a latent class or structured interpretation selected from a posterior distribution. Pragmatic Inference Failure and Illocutionary Force Mismatch therefore correspond to different errors in probabilistic reconstruction rather than to failures of physical transmission.

**Layer 7: Relational Interpretation as inference under relational priors**

Let $R$ denote the relational interpretation assigned to the message, including inferred motive, disposition, and stance:

$$R \sim p_7(r \mid I, K_L, H_{SL}, c),$$

where $H_{SL}$ represents relevant interaction history between speaker and listener.

Attributional Distortion can be represented as a biased prior or conditional model over relational states. For example,

$$p_L(R \mid I, c) \propto p_L(I \mid R, c)\, p_L(R \mid c).$$

A strongly biased prior over hostility, dismissiveness, competence, or intent may therefore produce a relational reconstruction that differs from the speaker's intended stance even when the linguistic content has been received accurately.

**Layer-resolved probabilistic factorization**

The seven forward analytical layers can now be expressed as a factorization rather than as a cascade of linear filters. One useful special case is as follows.

Let

$$\mathbf{Z} = \left(M, X, Y, A, I, R, \widehat{M}^{-}\right).$$

Then a useful factorization is

$$p(\mathbf{Z} \mid G, K_S, K_L, c) = q_1 q_3 q_4 q_5 q_6 q_7 q_D,$$

where

$$q_1 = p_1(M \mid G, W_S, c), \quad q_3 = p_3(X \mid M, K_S, c), \quad q_4 = p_4(Y \mid X, c_4),$$

and

$$q_5 = p_5(A \mid Y, K_L, c), \quad q_6 = p_6(I \mid A, K_L, c), \quad q_7 = p_7(R \mid I, K_L, H_{SL}, c), \quad q_D = p_D\left(\widehat{M}^{-} \mid A, I, R, K_L, c\right).$$

Layer 2 enters through $K_S$ and $K_L$ rather than as another value passed serially through the chain. This is consistent with the analytical interpretation of the layers: some are transformations of a communicated representation, while others are latent states that condition how those transformations are performed.

The factorization is intentionally representation-agnostic. The state variables may be categorical, symbolic, continuous, embedding-based, or learned latent representations. Likewise, the conditional models may be specified analytically, estimated statistically, or learned using modern machine-learning methods. The taxonomy therefore does not depend on any particular representation technology. Factorizing a joint distribution over a latent chain with observations, so that each variable is conditioned only on those it depends on, is the structure of a hidden Markov model (Rabiner, 1989).

### Connection to the definition of misunderstanding

Let the realized semantic or task distortion before repair be

$$D_c^- = d_c\big(M, \widehat{M}^-\big).$$

Using the unit-step function, the materiality criterion of Section 3.1 can be written as

$$\mu^- = u\big(D_c^- - \tau_c(M)\big),$$

where

$$u(z) = \begin{cases} 0, & z < 0, \\ 1, & z \geq 0. \end{cases}$$

Here $\tau_c(M)$ is the context-sensitive materiality threshold. A difference in reconstruction does not count as a material misunderstanding unless the resulting semantic or task loss reaches that threshold.

The probabilistic model also defines an episode-level risk of misunderstanding:

$$P_{\text{mis}}(c) = \Pr[D_c^- \geq \tau_c(M) \mid c].$$

This quantity differs from the realized binary state $\mu^-$. The first is a probability over possible episodes or uncertain states. The second records whether the threshold was crossed in a particular episode.

### Dialogue as a stochastic process

Dyadic communication unfolds over turns, so the layer-resolved model can be embedded in a state-space representation of the kind used for decision processes in which the state is not directly observed and actions are selected under a policy (Åström, 1965; Kaelbling et al., 1998), and which has been applied to dialogue systems in that form (Young et al., 2013). Let

$$S_t = \big(M_t, K_{S,t}, K_{L,t}, \widehat{M}_t, D_t, \mu_t\big)$$

represent the latent communication state at interaction time $t$. Let $U_t$ denote an interaction action, including continuation, acknowledgement, clarification, correction, or another repair move. The state evolves according to

$$S_{t+1} \sim p(S_{t+1} \mid S_t, U_t, c_t),$$

while the observable interaction record satisfies

$$O_t \sim p(O_t \mid S_t, c_t).$$

The latent state is not assumed to be directly observable to either participant or to an analyst. Observable evidence consists instead of utterances, timing, prosody, acknowledgements, corrections, actions, and other interactional signals. This distinction is particularly important for undetected misunderstandings, where the latent divergence exists without an explicit repair sequence revealing it.

### Layer 8: statistical detection and recovery

Layer 8 operates on evidence that the reconstructed meaning may differ from the intended one. Let

$$q_t = \Pr(\mu_t = 1 \mid O_{1:t})$$

be the posterior probability that a material misunderstanding exists given the observable interaction to time $t$. A simple statistical detector is

$$\delta_t = u(q_t - \gamma_t),$$

where $\gamma_t$ is the detection threshold. The distinction between $\tau_t$ and $\gamma_t$ is fundamental: $\tau_t$ determines whether a material misunderstanding exists, while $\gamma_t$ determines whether available evidence is sufficient to detect it.

A repair action $U_t$ produces additional evidence and therefore another posterior update. Schematically,

$$p_L(m \mid O_{1:t+1}) \propto p(O_{t+1} \mid m, O_{1:t}, U_t, c_t)\, p_L(m \mid O_{1:t}).$$

The resulting reconstruction becomes $\widehat{M}_{t+1}$, with post-repair distortion

$$D_{t+1} = d_c\left(M_t, \widehat{M}_{t+1}\right).$$

Repair is successful when

$$D_{t+1} < \tau_{c_{t+1}}(M_t),$$

or equivalently when the misunderstanding state returns to zero.

Recoverability can therefore be expressed as a probability of returning below the materiality threshold under an available repair policy. Over horizon $H$,

$$\mathcal{R}_H(t) = \max_{\pi} \Pr\left[D_{t+H} < \tau_{c_{t+H}}(M_t) \mid O_{1:t}, \pi\right],$$

where $\pi$ denotes a policy for selecting repair actions. This quantity remains separate from the definition of misunderstanding itself, consistent with Section 3.1.

### Mechanisms as perturbations of a nominal probabilistic model

The formal model separates the location of a mechanism from its effect. Let

$$p_j^{(0)}\left(z_j \mid \mathrm{pa}(z_j)\right)$$

be the nominal conditional distribution associated with Layer $j$, where $\mathrm{pa}(z_j)$ denotes the variables on which that state depends. Activation of mechanism $k$ may be represented as

$$p_j^{(k)}\left(z_j \mid \mathrm{pa}(z_j), \theta_k\right) \neq p_j^{(0)}\left(z_j \mid \mathrm{pa}(z_j)\right),$$

or as a perturbation to a conditioning state such as $K_S$ or $K_L$, or to the Layer-8 observation and repair policy.

This provides a common formal language for all eleven mechanisms without requiring them to share the same mathematical form. Encoding Ambiguity perturbs the formulation model $p_3(X \mid M, K_S, c)$; Channel and Medium Distortion perturbs $p_4(Y \mid X, c_4)$; Referential Misalignment and Presupposition Mismatch alter Layer-5 alignment; Pragmatic Inference Failure and Illocutionary Force Mismatch alter Layer-6 reconstruction; Attributional Distortion changes relational priors or likelihoods at Layer 7; and Repair Failure changes the probability that sufficient evidence is observed or acted upon at Layer 8.

The important consequence is that the taxonomy can be operationalized using statistical pattern recognition, probabilistic inference, or machine learning while retaining the same underlying theoretical structure.

The two conditioning states enter the factorization at different points, and that asymmetry is deliberate rather than notational. The sender-side state conditions formulation, because a speaker consults their model of the listener while composing. The receiver-side state conditions alignment, inference, and relational interpretation, because a listener arrives at an interpretation and adjusts it afterwards. This follows experimental evidence on both sides. Speakers plan without regard to common ground and then monitor and adjust, an adjustment that fails under time pressure (Horton & Keysar, 1996), and listeners interpret egocentrically by default and correct only with additional effort (Keysar et al., 2000). However, we separate the two states in order to represent an adjustment that can fail. Where audience design is integrated into planning rather than applied to it, the two coincide.

Writing the layers as a factorization gives the account two properties that a list of locations does not have, and both are adapted from the ladder of joint action (Clark, 1996). The first is downstream inheritance. Because each factor conditions only on its parents, perturbing any one of them changes the distribution of every variable downstream of it, so later layers may operate exactly as specified and still deliver a materially wrong reconstruction. The second is upstream evidence. Because the chain is a factorization, an observation that the reconstruction is materially correct is informative about every factor above it, which is why an acknowledgement at Layer 8 bears on the seven layers preceding it, and why Repair Failure can follow any mechanism rather than only those adjacent to it. The analogy is partial. Clark's levels are levels of description of one signalling act, related by presupposition, whereas these are factors in a chain related by conditioning, and Layer 2 sits outside the chain altogether, entering through the conditioning states rather than as a further factor.

## 4.3 Three Functional Roles: Generate, Amplify, Detect

In this section, we turn to a pattern that becomes visible once the eleven mechanisms are defined and their locations set out. They do not all do the same kind of work. Some create a divergence where there was none, some worsen one that already exists, and one governs whether any of it is ever seen. We distinguish three functional roles accordingly.

- Generate: Create a divergence between intended and reconstructed meaning.
- Amplify: Intensify, distort, or entrench an existing divergence.
- Detect: Govern whether the divergence becomes visible and whether it is repaired.
- These are theoretical assignments rather than measured frequencies. We assign each mechanism the role that follows from what it characteristically does to the system, and we make no claim to have measured how often any mechanism generates rather than amplifies across episodes. Establishing those proportions would require, however, the coded episode data that Section 6.6 identifies as the next step.

### Formal role definitions

Let $D^{(0)}$ denote the distortion under the corresponding nominal model and $D^{(k)}$ the distortion when mechanism $k$ is active. Let

$$\mu^{(0)} = u\left(D^{(0)} - \tau_c(M)\right)$$

and

$$\mu^{(k)} = u\left(D^{(k)} - \tau_c(M)\right).$$

A mechanism acts as a generator in an episode when its activation causes the first material threshold crossing:

$$\mu^{(0)} = 0 \quad \text{and} \quad \mu^{(k)} = 1.$$

Equivalently,

$$D^{(0)} < \tau_c(M) \quad \text{and} \quad D^{(k)} \geq \tau_c(M).$$

A mechanism acts as an amplifier when an existing divergence becomes larger or more consequential:

$$D^{(k)} > D^{(0)} > 0.$$

At the population level, amplification may also be expressed as an increase in misunderstanding risk:

$$P_{\text{mis}}^{(k)} > P_{\text{mis}}^{(0)}.$$

A mechanism affects detection and recovery when it changes the posterior probability of detecting an existing misunderstanding, the detector state $\delta_t$, or the probability of returning below threshold under repair. Repair Failure is therefore represented not by an increase in the original semantic divergence but by a degradation in observability, repair initiation, repair effectiveness, or recoverability:

$$\mathcal{R}_H^{(11)}(t) < \mathcal{R}_H^{(0)}(t),$$

or, in an undetected case,

$$\mu_t = 1 \quad \text{while} \quad \delta_t = 0.$$

These definitions preserve the distinction between the role a mechanism dominantly performs and the role it plays in a particular episode. That distinction matters for two of the eleven, and we set out both cases once the roles have been stated in ordinary language.

The GAD distinction therefore separates three different intervention problems. Prevention targets the mechanisms that generate divergence. Robustness targets the conditions that amplify or entrench it. Detection and repair target the observation and feedback process through which an existing divergence becomes visible and can be corrected.

We refer to these functions as generate, amplify, and detect, or GAD. Here, *detect* is shorthand for governing detection and repair. Of the eleven mechanisms, eight primarily generate divergences, two primarily amplify them, and one governs whether a divergence is detected and repaired.

These three functions are not our invention. Elements of the distinction appear separately in conversation analysis, which separates a trouble source from the repair practices through which it becomes visible, in the study of human error, and in root-cause analysis as practiced in healthcare and engineering. What this synthesis contributes, however, is their integration into a common vocabulary and their systematic assignment to specific mechanisms of misunderstanding, which Table 3 sets out.

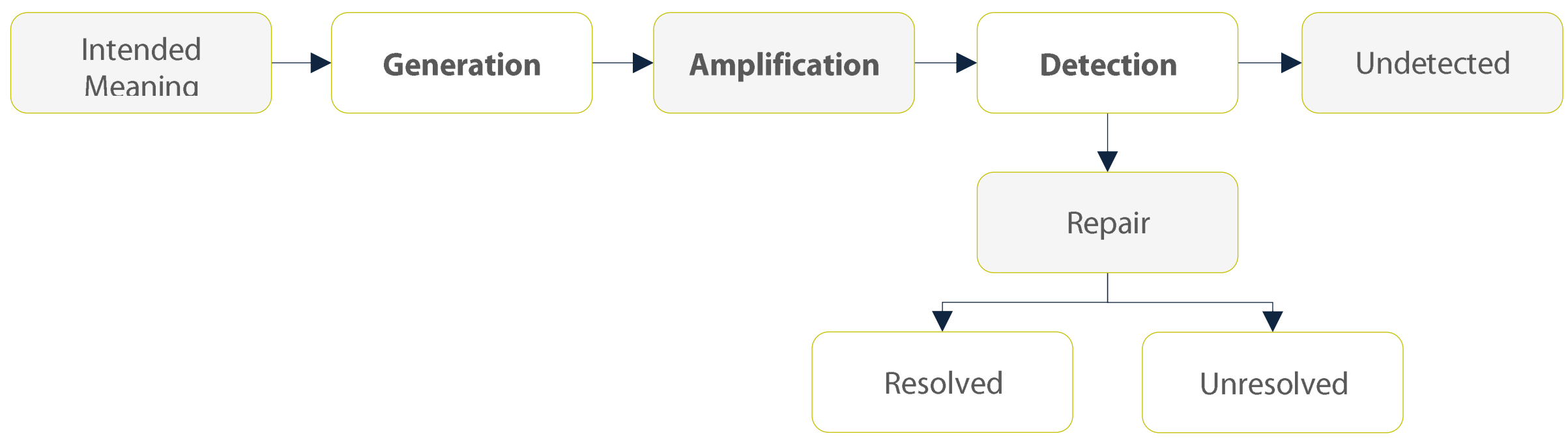

*Figure 2. The divergence-to-repair pathway. A divergence may be amplified, remain undetected, or become detected and either resolved or unresolved.*

Amplification is not required in every episode. A generated divergence may proceed directly to detection and repair, or it may remain undetected without having been substantially amplified.

The role assigned to a mechanism is its usual function rather than a fixed property, and two mechanisms in particular can act outside theirs. Channel and Medium Distortion normally amplifies another failure, by removing cues or opportunities for grounding and repair. However, it generates a divergence directly when the medium changes the content received, as with dropped audio, garbled transcription, a missing attachment, or a cropped image. Attributional Distortion normally amplifies a divergence, by turning a communication problem into a judgement about the other person. However, it can generate the initial divergence when the misunderstood object is the speaker's motive, intent, or relational stance, and it may also suppress detection by discouraging clarification.

The taxonomy therefore records each mechanism's dominant role and principal layer, whereas applied coding records the role it played in the episode being analysed. We provide the coding procedure in Appendix B.

*Table 2. The Eleven Mechanisms, Their Analytical Layers, and Dominant Functional Roles*

| DOMINANT ROLE | LAYER | MECHANISM |
|---|---|---|
| **1. GENERATE** | 1. Formation | 1. Intent Underspecification |
| **1. GENERATE** | 2. Perspective and Common Ground | 2. Common-Ground Overestimation |
| **1. GENERATE** | 2. Perspective and Common Ground | 3. Perspective-Taking Failure |
| **1. GENERATE** | 3. Formulation | 4. Encoding Ambiguity |
| **2. AMPLIFY** | 4. Medium and Interaction Constraints | 5. Channel and Medium Distortion |
| **1. GENERATE** | 5. Semantic and Pragmatic Alignment | 6. Referential Misalignment |
| **1. GENERATE** | 5. Semantic and Pragmatic Alignment | 7. Presupposition Mismatch |
| **1. GENERATE** | 6. Inference and Interpretation | 8. Pragmatic Inference Failure |
| **1. GENERATE** | 6. Inference and Interpretation | 9. Illocutionary Force Mismatch |
| **2. AMPLIFY** | 7. Relational Interpretation | 10. Attributional Distortion |
| **3. DETECT** | 8. Regulation and Recovery | 11. Repair Failure |

Three distinctions shaped the final taxonomy. Presupposition Mismatch was retained as separate from Common-Ground Overestimation, because one concerns a specific presupposed proposition while the other concerns a broader estimate of shared knowledge. Referential and presupposition failures were placed in an alignment layer, rather than treated as transmission failures. Channel and Medium Distortion and Attributional Distortion were classified primarily as amplifiers, because they usually worsen or entrench a divergence generated elsewhere.

## 4.4 Convergence Assessment

The convergence ratings reflect the number of independent research bodies supporting each mechanism, together with the range of evidence types represented. They assess the breadth of support for a mechanism, and not its frequency, effect size, or practical importance. We assigned the ratings using the rubric set out in Table 3.

*Table 3. Rubric for Assessing Cross-Disciplinary Convergence*

| RATING | THRESHOLD |
|---|---|
| **HIGH** | Supported by three or more independent research bodies, with both theoretical and empirical evidence |
| **MODERATE** | Supported by two research bodies, or by one research body with strong empirical evidence but limited uptake elsewhere |
| **EMERGING** | Supported mainly by recent work that has not yet developed cross-disciplinary support |
| **CONTESTED** | Defined differently across research bodies, or disputed as a mechanism rather than a symptom or outcome |

The resulting assessments are shown in Table 4.

*Table 4. Cross-Disciplinary Convergence Assessment for the Eleven Mechanisms*

| MECHANISM | RESEARCH BODIES REPRESENTED | EVIDENCE TYPES | CONVER-GENCE |
|---|---|---|---|
| **INTENT UNDERSPECIFICATION (M1)** | • Psycholinguistics Conversation Analysis<br>• Linguistic Typology<br>• Organisational Sensemaking | • Theoretical Experimental<br>• Corpus | High |
| **COMMON-GROUND OVERESTIMATION (M2)** | • Psycholinguistics<br>• Social Psychology<br>• Behavioural Economics<br>• Cognitive Science<br>• Organisational Ethnography | • Theoretical<br>• Experimental<br>• Ethnographic | High |
| **PERSPECTIVE-TAKING FAILURE (M3)** | • Cognitive Psychology<br>• Psycholinguistics<br>• Dialogue Theory<br>• Experimental Linguistics | • Theoretical<br>• Experimental | High |
| **ENCODING AMBIGUITY (M4)** | • Communication Theory<br>• Psycholinguistics<br>• Computational Dialogue | • Theoretical<br>• Experimental<br>• Corpus<br>• Computational | High |
| **CHANNEL AND MEDIUM DISTORTION (M5)** | • Organisational Communication<br>• Psycholinguistics<br>• CSCW<br>• Nonverbal Communication<br>• Workplace Studies | • Theoretical<br>• Experimental<br>• Review<br>• Ethnographic | High |
| **REFERENTIAL MISALIGNMENT (M6)** | • Psycholinguistics<br>• Conversation Analysis<br>• Computational Dialogue | • Theoretical<br>• Experimental<br>• Conversation-Analytic<br>• Corpus | High |
| **PRESUPPOSITION MISMATCH (M7)** | • Philosophy<br>• Formal Semantics<br>• Discourse Semantics<br>• Experimental Psycholinguistics | • Theoretical<br>• Formal-Semantic<br>• Experimental | High |
| **PRAGMATIC INFERENCE FAILURE (M8)** | • Pragmatics<br>• Relevance Theory<br>• Cognitive Linguistics | • Theoretical<br>• Experimental | High |

| MECHANISM | RESEARCH BODIES REPRESENTED | EVIDENCE TYPES | CONVER-GENCE |
|---|---|---|---|
| **ILLOCUTIONARY FORCE MISMATCH (M9)** | • Philosophy Of Language<br>• Pragmatics<br>• Intercultural Communication<br>• Conversation Analysis | • Theoretical<br>• Conversation-Analytic<br>• Applied | High |
| **ATTRIBUTIONAL DISTORTION (M10)** | • Social Psychology | • Theoretical<br>• Experimental | Moderate |
| **REPAIR FAILURE (M11)** | • Conversation Analysis<br>• Linguistic Typology<br>• Human-Machine Interaction<br>• Computational Linguistics | • Theoretical<br>• Experimental<br>• Conversation-Analytic<br>• Corpus<br>• Computational<br>• Ethnographic | High |

Ten of the eleven mechanisms receive a High rating. Attributional Distortion receives a Moderate rating, because its mechanism-specific support is concentrated in fewer research bodies.

Presupposition Mismatch receives High, because formal-semantic and discourse-semantic accounts are supplemented by experimental processing evidence (Chemla & Bott, 2013; Van Der Sandt, 1992). Channel and Medium Distortion also receives High, despite its dominant role as an amplifier. The convergence rating concerns support for the existence of the mechanism, and not the function it usually performs.

Appendix A provides the source-by-source evidence matrix underlying these assessments.

# 5. Applying the Taxonomy to Dialogue Cases

In this section, we apply the taxonomy to nine cases. Three are drawn from documented material, being a corpus conversation in Case 1, an experimental episode in Case 7, and an accident investigation in Case 8. Case 3 is a stylised clinical vignette based on documented handover failures, and the remaining five are constructed illustrations.

The cases demonstrate how the taxonomy can be applied, but they do not validate it. Validation would require independent coders applying the coding manual to a shared set of cases, and reporting their agreement.

## 5.1 Coding Approach

For each case, we identify the misunderstood object, the primary mechanism, any contributing mechanisms, and the final detection and resolution state.

The primary mechanism is the mechanism that produced the divergence, and removing it would most directly have prevented the misunderstanding. A contributing mechanism enabled, amplified, or concealed the divergence, but it did not independently produce it.

This distinction prevents every plausible mechanism from being assigned equal causal importance.

## 5.2 Case 1: Referential Misalignment

**Source.** A documented conversation, originally from the AudioBNC corpus, and analysed in earlier work (Albert & de Ruiter, 2018)**.**

**Episode.** A mother asks Claire to "switch that off," referring to the television. Claire interprets "that" as referring to the oven alarm.

**Analysis.** The primary mechanism is Referential Misalignment (M6). The speaker and listener connect the same expression to different objects. The mother corrects the referent in the following turn.

**State.** Detected and resolved. Repair succeeds promptly, so Repair Failure is not assigned as a contributing mechanism.

## 5.3 Case 2: Misunderstanding an Implied Refusal

**Source.** A constructed illustration based on (Grice, 1975) pragmatics and indirect refusals.

**Episode.** In response to an invitation, the speaker says, "I've got a class at noon," intending a polite refusal. The listener interprets the statement literally and assumes the invitation remains open.

**Analysis.** The primary mechanism is Pragmatic Inference Failure (M8) because the listener does not recover the implied refusal. Illocutionary Force Mismatch (M9) contributes because the utterance is interpreted as an informational statement rather than a refusal. Repair Failure (M11) allows the divergence to remain unnoticed.

**State.** Undetected.

## 5.4 Case 3: Clinical Handover

**Source.** A stylised clinical vignette based on documented patterns in healthcare handover failures (Haig et al., 2006).

**Episode.** Clinician A reports that a patient has "some abdominal discomfort." A has in mind moderate pain and a low red blood cell count suggesting internal bleeding. Clinician B interprets the statement as describing routine postoperative discomfort.

**Analysis.** The primary mechanism is Common-Ground Overestimation (M2). Clinician A assumes that Clinician B shares the broader clinical picture and therefore does not state the suspected bleeding. Encoding Ambiguity (M4) contributes because "some" does not specify severity. Repair Failure (M11) contributes because neither clinician checks the interpretation.

**State.** Undetected.

## 5.5 Case 4: Cross-Cultural Illocutionary Force Mismatch

**Source.** A constructed illustration based on cross-cultural pragmatics and indirect refusal conventions (House, 2006).

**Episode.** In a cross-cultural workplace exchange, one manager says, "That would be very difficult," intending a polite refusal. A colleague accustomed to more direct refusals interprets the statement as a request for help overcoming the difficulty.

**Analysis.** The primary mechanism is Illocutionary Force Mismatch (M9) because a refusal is interpreted as a request. Pragmatic Inference Failure (M8) contributes because the implied refusal is not recovered. Repair Failure (M11) allows the different interpretations to remain hidden. Case 2 is the mirror of this one. Both are indirect refusals that are not taken as refusals, and both implicate the same two mechanisms with the assignments reversed. There the utterance is treated as information and no act is recovered at all, so Pragmatic Inference Failure is primary. Here an act is recovered and it is the wrong one, so Illocutionary Force Mismatch is primary. The contrast is what shows the two mechanisms to be separable.

**State.** Undetected.

## 5.6 Case 5: Attributional Distortion as a Generator

**Source.** A constructed illustration based on attribution research and computer-mediated communication (Heider, 1958; Ross, 1977).

**Episode.** Party A sends a brief text response to Party B's detailed analysis. Party B interprets the brevity as dismissiveness. Party A was under time pressure and considered the response complete.

**Analysis.** The primary mechanism is Attributional Distortion (M10). Party B interprets situational behaviour as evidence of Party A's attitude or character. In this case, M10 generates the divergence because the misunderstood object is relational stance rather than message content.

Channel and Medium Distortion (M5) contributes because text removes tone and other cues that might have indicated time pressure. Repair Failure (M11) contributes because Party B does not seek clarification.

**State.** Undetected.

## 5.7 Case 6: Presupposition Mismatch

**Source.** A constructed illustration based on (Stalnaker, 1973, 2002) account of presupposition and accommodation.

Episode. Party A asks Party B, "Why did you stop sending the weekly report?" Party A believes that Party B discontinued the reports. Party B replies, "I have not stopped sending them; the distribution list may have changed." Party A insists that no report was received, but the exchange ends without establishing whether the reports were sent, delivered, or routed correctly.

Analysis. The primary mechanism is Presupposition Mismatch (M7). Party A's question presupposes that Party B stopped sending the reports, while Party B rejects that proposition. Common-Ground Overestimation (M2) contributes because Party A assumes that the alleged interruption is mutually recognised. Repair Failure (M11) contributes because the clarification exposes the divergence but does not resolve the underlying factual disagreement.

The case is classified primarily as Presupposition Mismatch rather than Common-Ground Overestimation because the divergence is carried by the specific linguistic trigger "stop," which presupposes a previous activity and its termination. Common-Ground Overestimation is contributing because Party A assumes that this proposition is mutually recognised.

State. Detected but unresolved.

## 5.8 Case 7: Perspective-Taking Failure

**Source.** A documented experimental episode from the director task (Keysar et al., 2000).

**Episode**. A director instructs a participant to move the tape, referring to a cassette tape visible to both. A roll of adhesive tape occupies a slot that is occluded from the director and visible only to the participant. The participant looks first towards the occluded slot, and reaches for the adhesive tape rather than the cassette.

**Analysis**. The primary mechanism is Perspective-Taking Failure (M3). The information needed to take the director's perspective is available and visibly marked, since the occluding backs of the hidden slots face the participant, and the listener interprets from their own vantage point regardless. Common-Ground Overestimation (M2) does not apply, because the listener holds an accurate model of what the director can see and fails to apply it rather than holding a mistaken one. The episode is representative rather than singular.

Across the study, participants looked longer at the occluded slot when it held the competitor object than when it held an unrelated one, and a substantial proportion reached for it.

**State.** Detected and resolved in the experimental setting, where the task reveals the error immediately. In ordinary conversation the equivalent divergence need not surface.

## 5.9 Case 8: Encoding Ambiguity

**Source.** A documented aviation accident, analysed linguistically (Cushing, 1994)**.**

**Episode.** In fog at Tenerife on 27 March 1977, the captain of a departing aircraft transmits that they are now at takeoff. He means that the aircraft has begun its takeoff roll. The controller understands a position report, that the aircraft is holding at the point of takeoff. The aircraft collides with another on the runway and 583 people are killed.

**Analysis.** The primary mechanism is Encoding Ambiguity (M4). The phrase permits both a progressive reading, that the roll has commenced, and a locative reading, that the aircraft stands at the takeoff position, and the listener selects the reading that was not intended. Channel and Medium Distortion (M5) contributes, because a radio channel without visual contact removes the confirmation that fog had already denied. Repair Failure (M11) contributes, because the exchange that followed restored conversational progress without resolving which reading had been received.

**State.** Undetected. The divergence is never surfaced within the episode.

## 5.10 Case 9: Intent Underspecification

**Source.** A constructed illustration, built to satisfy the observability criterion set out in Section 4.1 (Brown-Schmidt & Konopka, 2015).

**Episode.** A manager asks a colleague to look into the Henderson account before Friday. The colleague prepares a written risk analysis. The manager replies that this is not what was needed but is unable to say what was, and over two further exchanges the request is reformulated, first as a call to the client and then as a short summary for a meeting.

**Analysis.** The primary mechanism is Intent Underspecification (M1). The successive reformulations, and the inability to state what was wanted when asked directly, indicate that the intended meaning was not determinate at the moment the request was made. That is the test which separates this mechanism from Encoding Ambiguity (M4), where the speaker holds a settled intention and expresses it in a form that permits another reading. Here no single wording would have served, because there was not yet a message to word.

**State.** Detected and resolved, after two cycles of repair. Repair Failure (M11) is not assigned, because the checking that the taxonomy places at Regulation and Recovery occurred and eventually succeeded.

## 5.11 Comparison of the Cases

The cases show how the same episode may involve several mechanisms without assigning them equal causal weight. The primary mechanism identifies what produced the divergence, whereas contributing mechanisms explain the conditions that enabled, amplified, or concealed it. Across the nine cases, every mechanism whose dominant role is to generate a divergence appears once as a primary mechanism. Channel and Medium Distortion and Repair Failure appear only as contributing mechanisms, which follows from the definition of a primary mechanism rather than from a gap in the selection, since those two amplify an existing divergence and govern whether it is detected.

*Table 6. Comparison of the Nine Dialogue Cases by Misunderstood Object, Mechanisms, and Final State*

| CASE | MISUNDERSTOOD OBJECT | PRIMARY MECHANISM | CONTRIBUTING MECHANISMS | FINAL STATE |
|---|---|---|---|---|
| **1** | Referent | M6: Referential Misalignment | None | Detected and resolved |
| **2** | Implied meaning and communicative act | M8: Pragmatic Inference Failure | M9, M11 | Undetected |
| **3** | Clinical severity and urgency | M2: Common-Ground Overestimation | M4, M11 | Undetected |
| **4** | Communicative act | M9: Illocutionary Force Mismatch | M8, M11 | Undetected |
| **5** | Relational stance | M10: Attributional Distortion | M5, M11 | Undetected |
| **6** | Presupposed proposition | M7: Presupposition Mismatch | M2, M11 | Detected but unresolved |
| **7** | The other participant's visual perspective | M3: Perspective-Taking Failure | None | Detected and resolved |
| **8** | Requested action and aircraft state | M4: Encoding Ambiguity | M5, M11 | Undetected |
| **9** | The requested action itself | M1: Intent Underspecification | None | Detected and resolved |

# 6. Discussion

## 6.1 Areas of Convergence

Ten of the eleven mechanisms receive High convergence ratings. The broadest support concerns failures of alignment and common ground, being Referential Misalignment (M6), Presupposition Mismatch (M7), Common-Ground Overestimation (M2), and Perspective-Taking Failure (M3). These mechanisms are related, but they should remain separate. They concern, respectively, identifying the same referent, sharing a particular presupposition, estimating what background knowledge is shared, and applying the other person's perspective during communication.

Pragmatic Inference Failure (M8) and Illocutionary Force Mismatch (M9) also receive strong support. Both become more likely when participants rely on different cultural conventions. However, intercultural difference is better treated as a condition affecting existing mechanisms than as a separate mechanism in its own right.

Repair Failure (M11) differs from the other mechanisms, because it can follow any of them, which follows from the composition of the layers set out in Section 4.2. The organization of repair is largely independent of the source of the trouble (Schegloff, 1987), and repair practices recur frequently across languages (Dingemanse et al., 2015). Repair Failure therefore determines whether a divergence is exposed and addressed, or whether it remains undetected.

One mechanism receives a Moderate rating rather than High. Attributional Distortion (M10) rests on a well-established experimental literature, but that literature sits almost entirely within social psychology. The convergence rating measures breadth across research bodies rather than the strength of evidence within any one of them. Notably, the rating does not imply that the mechanism is weakly evidenced. It implies that the evidence has not yet arrived from independent directions.

Intent Underspecification (M1) is the one rating that moved during revision. It was rated Moderate until the timing of conversational turns was recognised as a third research body bearing on it. That evidence establishes the condition under which a speaker commits before the intended meaning has settled, rather than demonstrating a divergence in any particular episode, so the mechanism still has to be inferred from later behaviour when a case is coded. Section 6.6 records how that third body came to be searched for.

## 6.2 The Generate, Amplify, and Detect Distinction

The distinction among generation, amplification, and detection has precedents in several fields. The separation of a trouble source from the practices that expose it, which Section 3.1 used to hold detection apart from the existence of a divergence, is itself one of them (Schegloff, 1987). A comparable structure appears in the account of active failures, latent conditions, and detection barriers (Reason, 1990), and again in root-cause analysis as practised in healthcare and engineering.

We did not import this distinction and apply it to the mechanisms. The typing emerged from the mechanisms themselves, and it then proved to reproduce a distinction that three unrelated literatures had each reached on their own evidence. We read that convergence as a check rather than as a debt. A classification assembled from nine fields that arrives at a distinction conversation analysis, human-error theory, and root-cause analysis reached separately is a classification tracking a real difference. What the GAD framework adds is a common vocabulary for the three functions and their systematic assignment to specific mechanisms of misunderstanding. The distinction separates the process that creates a divergence from the conditions that worsen it, and from the practices that expose and repair it.

The separation carries practical implications that a single undifferentiated notion of communication quality obscures. Improvements to message formation and interpretation reduce the rate at which divergences are generated, and this is where most communication training is currently directed. However, such improvements will not prevent a divergence from persisting once it has been created, because persistence is governed by the adequacy of repair rather than by the quality of the original formulation. Conversely, stronger feedback and confirmation practices may raise the detection rate substantially without reducing the mechanisms that repeatedly generate divergences in the first place. Prevention and detection therefore address different parts of the problem, and an intervention aimed at one should not be credited with the benefits of the other.

## 6.3 What the Cases Show

Section 5 applies the taxonomy to nine episodes, and what that exercise tests is not whether a mechanism can be named but whether a competing one can be ruled out. In several of the cases the assignment turns on an exclusion, and each exclusion rests on something observable in the episode rather than on the judgement of the person coding it. Perspective-Taking Failure is separated from Common-Ground Overestimation in the director task by whether the listener holds a mistaken model of what the speaker can see, or holds an accurate one and fails to apply it. Intent Underspecification is separated from Encoding Ambiguity by whether the intention was settled at the moment of speaking, which the successive reformulations answer. Repair Failure is withheld in two cases because the checking that the taxonomy places at Regulation and Recovery occurred and succeeded. A taxonomy is usable to the degree that its exclusions can be stated with a reason, and that is the property the nine cases exercise.

The final state varies across the cases and is recorded separately from the mechanism that produced the divergence. Five of the nine remain undetected, one is detected but unresolved, and three are detected and resolved. Recording the two independently is what the functional roles require, because which mechanism produced a divergence and whether the interaction caught it are governed by different layers, a point Section 6.2 argues on other grounds. What the nine cases cannot show is how often each state occurs, since they were selected to cover the mechanisms rather than sampled from a population of episodes, and Section 6.6 sets out what establishing that would require.

## 6.4 Implications for AI-Mediated Communication

Section 1.1 sets out why these mechanisms matter now, that communication is moving into channels which withdraw the resources repair depends on. What a mechanism-level account adds to that argument is the ability to say which layer a particular AI agent removes. An AI agent that summarises withdraws grounding resources at Medium and Interaction Constraints while leaving Formulation intact. An AI agent that composes on a communicator's behalf acts at Formulation, under a conditioning state the communicator never inspects. An AI agent that relays between two others removes Regulation and Recovery altogether, because no party in the chain is positioned to notice a divergence introduced earlier. Those are three different failures, and an evaluation reporting only whether an exchange satisfied its participants would score them alike.

The distinction bears on where design effort is spent. Improving the fluency of generated messages addresses the layers at which divergences are generated, and leaves detection where it was. Building confirmation and read-back into an interface addresses detection, and leaves generation where it was. Section 6.2 argues that the two are not substitutes, and the argument carries more weight here, because a single AI agent can improve one while withdrawing the other. A channel of that kind would generate fewer divergences and catch a smaller share of the ones it still generates, and it would report as an improvement on any measure that does not separate the two. The evidence reviewed in this paper is evidence about people, so these are implications drawn from the taxonomy rather than findings about deployed systems, and testing them belongs with the coded episode data called for in Section 6.6.

## 6.5 Gaps Revealed by the Synthesis

In this section, we set out the gaps the synthesis reveals, commencing with the point at which its own distinctions are least secure. Five of the eight layers carry a single mechanism and three carry two. The three carrying two are the points at which the reviewed literatures maintain a distinction that does not survive being collapsed, being reference against presupposition, implied meaning against illocutionary act, and a mistaken model of the other participant against a failure to apply a correct one. That is where the taxonomic work of this synthesis is concentrated, and it is therefore where independent coders are most likely to disagree. We expect any revision of the taxonomy to reach those three layers first.

Interactive alignment is underrepresented in current accounts of misunderstanding. Research often emphasises deliberate perspective-taking and common-ground reasoning, while giving less attention to the automatic convergence of words, structures, and representations that interactive alignment describes (Pickering & Garrod, 2004). Failures of this automatic alignment may currently be classified under Common-Ground Overestimation or Perspective-Taking Failure, even where neither fully captures the process.

Audience or recipient design is also underdeveloped. Speakers may fail to adapt vocabulary, form, or level of detail to what a listener is known to understand (Bell, 1984; Clark & Murphy, 1982). Such production-side failures are harder to identify than listener interpretations because the intended audience design must often be inferred from later behaviour.

Computational work has begun to identify self-repair and other-repair in dialogue, but existing models generally focus on limited repair types and rely on domain-specific knowledge (Purver et al., 2018). Extending this work to the broader taxonomy would support empirical validation and may help identify misunderstandings that participants themselves do not detect.

Regulation and Recovery is also the least decomposed of the eight layers. It carries a single mechanism, while a model of failure handling in human-robot interaction separates how a failure is communicated from how it is perceived and understood and from how it is solved (Honig & Oron-Gilad, 2018). Our three detection and resolution states map onto that sequence closely enough that the layer could plausibly be divided along the same lines, which would let a repair that is initiated but not understood be distinguished from one that is understood but not acted on. We have not made that division here, because it would restructure the taxonomy rather than extend it, and we note it as the most promising direction for refining the layer.

Section 4.2 supplies the bridge to computational operationalisation that the semantic-distance formulation of Section 3.1 only gestured at, by writing the layers as conditional distributions and the reconstruction as an inference under loss. The materiality threshold is no longer a free parameter, since it can be read as the smallest semantic distortion sufficient to change the task-relevant consequence. The distance function itself remains unspecified, however, and future work should instantiate and validate it for particular misunderstood objects, like referents, clinical severity, requested actions, or relational stance. We regard a formal treatment of the individual mechanisms, expressed in the same notation, as the natural next step, and we develop it separately rather than here.

Two things follow from that formulation which the taxonomy alone could not state. The first concerns detection. Writing the state of an episode as latent, and the evidence available to either participant as an observation of it, separates whether a material divergence exists from whether the interaction has produced enough evidence to reveal one. The undetected misunderstanding, which is the case this paper treats as most consequential and least visible, is then a single expression rather than a description: the divergence has crossed the threshold while the detector has not fired. The second concerns where a mechanism acts. Because the layers compose, a mechanism can be located as a perturbation of one conditional distribution while its effects reach every layer downstream, which is what allows the taxonomy to place a mechanism at one layer without confining its consequences there.

What the formulation does not supply is equally worth stating. No prior is specified, no inference procedure is given, and nothing is estimated from data. The model is a way of writing the taxonomy down rather than a system that has been fitted, and the work of instantiating it belongs with the mechanism-level treatment we have deferred.

The evidence base is also uneven across settings. Legal, diplomatic, crisis, organizational, and workplace communication may produce different combinations of mechanisms from those found in laboratory, healthcare, and interpersonal studies. Workplace research already contains useful naturalistic material. One ethnography of technicians, for example, showed how operational knowledge circulated through informal stories, and how failures in that circulation led to misunderstanding and repeated problem solving (Orr, 1996). Comparable material remains underused in research on misunderstanding (Heath & Luff, 2000; Suchman, 1987).

## 6.6 Limitations

The nine fields were selected because each offers a distinct account of how misunderstanding arises, and the boundaries between them are ours. Notably, mechanisms concentrated in fields outside those nine may therefore have been missed, and the synthesis cannot rule that out.

A convergence rating is therefore a property of this synthesis as much as of the literature. It records how widely we found a mechanism supported, not how widely it has been studied, and the two come apart whenever a mechanism is addressed in a field we did not search for it. A targeted search for one further research body can raise a rating without anything in the literature having changed. The ratings should be read as a lower bound on cross-disciplinary convergence and as a map of where attention has demonstrably met, rather than as a measure of how well established a mechanism is. The corollary matters for anyone extending this work: a mechanism rated Moderate is an invitation to look in a field we did not.

One rating in this paper moved on exactly that basis, and the standard just stated requires us to say so. Intent Underspecification was rated Moderate, and its supporting evidence was then found to rest on a source that did not establish what it was cited for. Searching for better support led to the literature on the timing of conversational turns, which is a third research body, and the rating rose to High. Nothing in the literature changed. What changed is that we looked in a field we had not previously searched for that mechanism, having been given a reason to look. We report it because a rating that moves this way is a finding about the reach of our search before it is a finding about the reach of the evidence.

The convergence ratings retain an element of interpretive judgement. Appendix A makes the source assignments, research-body counts, and evidence types visible, but it cannot eliminate judgement in deciding whether a source supports a particular mechanism. Independent coding is needed to test those assignments and the resulting ratings. There is a precedent for how to do this in a directly comparable case. An error taxonomy for dialogue systems was built by integrating a theory-driven scheme with a data-driven one, which is the problem this synthesis faces across nine fields rather than one, and was then evaluated by having multiple annotators code the same material and reporting weighted agreement against both schemes it was built from (Higashinaka et al., 2021). That design transfers directly. It would test not only whether coders agree, but whether the consolidation improved on the accounts it consolidated.

The literature is also affected by English-language and WEIRD-sample bias. Some mechanisms may recur widely across languages and cultures, but their frequency, expression, and consequences are likely to vary. The cross-language evidence for repair supports broad recurrence of repair practices, but it does not establish that all mechanisms operate identically across settings.

A further limitation is that evidence about misunderstanding often becomes available only after repair. Researchers can usually identify a divergence because someone later corrects, questions, or clarifies it. This creates a bias towards detected misunderstandings, even though undetected cases may be more consequential. Developing methods for identifying divergences that never produce an observable repair sequence remains a central methodological challenge. The earlier literature already names this case. Alongside misunderstandings recognised at once and those recognised only later, a third category covers those the participants never recognise although an analyst can see them in the record (Linell, 1995). Our undetected state and that latent category pick out the same episodes, and adopting the term makes the methodological requirement precise: identifying them calls for a record and an independent standard of what was meant, rather than for anything the participants themselves supply.

Finally, the dialogue cases illustrate rather than validate the taxonomy. Three of the nine come from documented material, being a corpus conversation, an experimental episode, and an accident investigation. The remaining six are constructed or stylised, and all nine were coded by one author. Intent Underspecification is deliberately among the constructed ones. A documented case would require evidence that the intended meaning was not determinate at the moment of speaking, and Section 4.1 sets out why that is available only by inference from later behaviour. The instances the literature does record are self-repairs, in which the speaker recovers before anything reaches the listener, so they document the mechanism without documenting a misunderstanding. Validation requires independent coders applying the manual in Appendix B to a shared collection of naturally occurring cases, and reporting agreement. Relevant sources include the AudioBNC and CABNC corpora, the HCRC MapTask corpus, and the cross-language repair corpus used by (Dingemanse et al., 2015).

# 7. Conclusion

We have integrated research from nine fields into a mechanism-level account of misunderstanding in two-party communication. From that synthesis, we derived a working definition of misunderstanding, and we identified eleven mechanisms organized into eight analytical layers and three functional roles, being generate, amplify, and detect. Ten mechanisms receive High convergence ratings, and Attributional Distortion receives Moderate support. The roles were not imported from an existing scheme. They emerged from the mechanisms themselves, and then proved to reproduce a distinction that conversation analysis, human-error theory, and root-cause analysis had each arrived at on separate evidence. No prior classification of misunderstanding both locates mechanisms at points in the communicative process and types them by the function they perform.

The central finding is that the literature does not point to a single cause of misunderstanding. It points to a process. Failures in forming and expressing meaning, in aligning reference and assumptions, in drawing inferences, and in using another person's perspective can all generate a divergence. Channel and attributional effects may then amplify it, whereas repair determines whether it becomes visible and is resolved.

The taxonomy contributes a shared vocabulary for distinguishing where a misunderstanding arises, what function each mechanism performs, and why some divergences persist. The evidence matrix in Appendix A makes the convergence assessments auditable, the coding manual in Appendix B provides an initial procedure for applying the taxonomy to cases, and the nine dialogue cases in Section 5 show that procedure in use. What those cases exercise is not whether a mechanism can be named but whether a competing one can be ruled out, and every exclusion in them rests on something observable in the episode. Writing the layers as a chain of conditional distributions adds one thing the vocabulary alone could not supply, which is a way of separating whether a material divergence exists from whether the interaction has produced enough evidence to reveal one. That separation is what makes the undetected case statable rather than only describable.

The framework remains conceptual rather than validated. The next step is to have independent coders apply the taxonomy to naturally occurring dialogue, and to assess whether the mechanisms can be identified reliably, distinguished from one another, and used across languages, cultures, media, and applied settings. Particular attention should be given to undetected misunderstandings, which may be the most consequential but are also the least visible in existing evidence. That attention is becoming harder to postpone. The conditions that have always made misunderstanding survivable, being continuous, cheap, and largely automatic repair, are being removed from an increasing share of human communication, and they are being removed faster than the means of detecting divergence are being built. The AI agents now composing, summarising, and relaying messages on people's behalf withdraw those conditions layer by layer, and an agent that improves the fluency of what it writes does not thereby improve the chance that anyone notices when it has misread. Learning to detect misunderstanding systematically is among the more useful kindnesses a field can do for the people who depend on its systems.

# Appendix A: Evidence Matrix

This appendix provides the source-by-source evidence underlying the convergence ratings in Section 4.4. The ratings were assigned using the rubric presented in that section, based on the number of independent research bodies supporting each mechanism and the range of evidence types represented.

The final block lists sources that inform the framing, the scope, the motivation, or the method rather than the rating of any single mechanism; these are marked as not counting towards a mechanism rating. A source that supports more than one mechanism appears under its primary one, with the secondary noted in the cell. The matrix makes the field and evidence-type counts auditable; it does not remove the interpretive step of assigning a source to a mechanism, which is what independent coding (Appendix B) is meant to test.

*Table 5. Source-by-Source Evidence Matrix Supporting the Mechanism Roles and Convergence Ratings*

| SOURCE | MECHANISM & ROLE | FIELD | EVIDENCE TYPE | BASIS FOR INCLUSION |
|---|---|---|---|---|
| **GILOVICH, SAVITSKY, & MEDVEC (1998)** | • M2, M1<br>• Generate | • Social Psychology | • Experimental | Illusion of transparency: speakers overestimate how evident their own internal states are to others |
| **WEICK (1995)** | • M1<br>• Generate | • Organisational Sensemaking | • Theoretical | Meaning and intention develop through interpretation rather than being fully formed in advance |
| **BROWN-SCHMIDT & KONOPKA (2015)** | • M1<br>• Generate | • Psycholinguistics | • Experimental | Message planning is continuously incremental, so speakers begin without a fully specified message and add elements after articulation starts |
| **LEVELT (1983)** | • M1<br>• Generate | • Psycholinguistics | • Experimental | Self-repair analysis: speakers interrupt an utterance in progress to replace the message being expressed, not only to correct its expression |
| **STIVERS ET AL. (2009)** | • M1<br>• Generate | • Conversation Analysis<br>• Linguistic Typology | • Corpus (10 Languages) | Median gaps between turns of 0 to 300 ms across ten languages, far shorter than the time required to plan an utterance |
| **HOLLER, KENDRICK, CASILLAS, & LEVINSON (2015)** | • M1<br>• Generate | • Psycholinguistics | • Theoretical | Producing even a one-word utterance takes at least 600 ms, so speakers plan while listening and launch before the message is fully specified |
| **CLARK & SCHAEFER (1989)** | • M2<br>• Generate | • Psycholinguistics | • Theoretical | The grounding mechanisms whose failure produces lopsided common ground |
| **ROSS, GREENE, & HOUSE (1977)** | • M2, M10<br>• Generate (M2), Amplify (M10) | • Social Psychology | • Experimental | False consensus effect |
| **CAMERER, LOEWENSTEIN, & WEBER (1989)** | • M2<br>• Generate | • Behavioural Economics<br>• Cognition | • Experimental | Curse of knowledge |
| **NEWTON (1990)** | • M2<br>• Generate | • Cognitive Psychology | • Experimental | Tapper study demonstrating overestimated transmission |
| **ROZENBLIT & KEIL (2002)** | • M2<br>• Generate | • Cognitive Science | • Experimental | Illusion of explanatory depth |
| **ORR (1996)** | • M2<br>• Generate (Context) | • Organisational Ethnography | • Ethnographic | Operational common ground carried informally by narrative |
| **KEYSAR, BARR, BALIN, & BRAUNER (2000)** | • M3<br>• Generate | • Cognitive Psychology | • Experimental | Egocentric interpretation is the fast default |
| **PICKERING & GARROD (2004)** | • M3<br>• Generate | • Psycholinguistics | • Theoretical | Interactive alignment as a coordination route separate from M2 |
| **ANDERSON & DILLON (2023)** | • M3<br>• Generate | • Psycholinguistics | • Experimental<br>• Corpus | Egocentric bias appears in grammatical perspective-taking, particularly in production |
| **SHANNON (1948)** | • M4<br>• Generate | • Information And Communication Theory | • Theoretical | Encoding fidelity is framed as the technical-level communication problem |
| **WEAVER (1949)** | • M4<br>• Generate | • Communication Theory | • Theoretical | Semantic noise is the conceptual ancestor of encoding ambiguity |
| **PAXTON, ROCHE, IBARRA, & TANENHAUS (2021)** | • M4, M5<br>• Generate | • Psycholinguistics<br>• Corpus | • Corpus<br>• Experimental | Utterance ambiguity and low lexical density predict miscommunication turn by turn |
| **DAFT & LENGEL (1984, 1986)** | •<br>• Amplify | • Organisational Communication | • Theoretical | Lean media lack the cues an ambiguous message needs |
| **CLARK & BRENNAN (1991)** | • M5, M2<br>• Amplify (M5), Generate (M2) | • Psycholinguistics<br>• CSCW | • Theoretical | Each medium removes specific grounding resources |
| **PATTERSON ET AL. (2023)** | • M5<br>• Amplify | • Nonverbal Communication | • Review | Nonverbal cue loss is not merely the loss of redundant information |
| **HEATH & LUFF (2000)** | • M5<br>• Amplify (Can Generate) | • CSCW<br>• Workplace Studies | • Ethnographic | A medium can strip away the interactional resources people coordinate through |
| **BEATTIE & SHOVELTON (1999)** | • M5<br>• Amplify | • Nonverbal Communication<br>• Gesture Research | • Experimental | Listeners who saw a speaker's iconic gestures recovered object position and size more accurately than listeners who heard the same speech alone, so a nonverbal channel carries information the words do not. Significant only for particular semantic categories |
| **SNEDEKER & TRUESWELL (2003)** | • M5<br>• Amplify | • Psycholinguistics | • Experimental | Speakers produce prosodic boundaries that let listeners resolve a syntactic ambiguity before the phrase completes. Prosody is absent from text entirely |
| **CLARK & WILKES-GIBBS (1986)** | • M6<br>• Generate | • Psycholinguistics | • Experimental | Reference is settled through an iterative, collaborative process |
| **SCHEGLOFF (1987)** | • M6, M9, M11<br>• Generate (M6, M9), Detect (M11) | • Conversation Analysis | • Conversation-Analytic | Problematic reference documented as a source of misunderstanding in real talk |
| **LI, GATT, & POESIO (2026)** | • M6<br>• Generate | • Computational Dialogue<br>• Corpus | • Corpus (LLM-Assisted) | Apparent grounding can mask referential misalignment; treated as suggestive |
| **STALNAKER (1973, 2002)** | • M7<br>• Generate | • Philosophy Of Language<br>• Formal Semantics | • Theoretical<br>• Formal | Common ground as a context set; presupposition |
| **LEWIS (1979)** | • M7<br>• Generate | • Formal Semantics | • Theoretical | Accommodation of an unshared presupposition |
| **CHEMLA & BOTT (2013)** | • M7<br>• Generate | • Experimental Psycholinguistics | • Experimental | Presuppositions are processed online, and accommodation carries a measurable cost; this is the experimental support that lifts M7 to High |
| **GRICE (1975)** | • M8<br>• Generate | • Pragmatics | • Theoretical | Cooperative principle and conversational implicature |
| **SPERBER & WILSON (1986/1995)** | • M8<br>• Generate | • Relevance Theory | • Theoretical | Relevance-guided selection of an interpretation |
| **GIORA (1997, 2003)** | • M8<br>• Generate | • Cognitive Linguistics | • Theoretical<br>• Experimental | Graded salience mismatch between speaker and listener |
| **AUSTIN (1962)** | • M9<br>• Generate | • Philosophy Of Language | • Theoretical | Locutionary, illocutionary, and perlocutionary acts |
| **SEARLE (1969)** | • M9<br>• Generate | • Philosophy Of Language | • Theoretical | Illocutionary force |
| **BROWN & LEVINSON (1987)** | • M9<br>• Generate | • Politeness<br>• Pragmatics | • Theoretical | Indirection scaled to social distance, power, and imposition |
| **THOMAS (1983)** | • M9<br>• Generate<br>• (Also | • Cross-Cultural Pragmatics | • Theoretical | Pragmatic failure; pragmalinguistic versus sociopragmatic |
| | • M8) | | | |
| **(LEECH, 1983)** | • M9<br>• Generate | • Pragmatics | • Theoretical | Source of the pragmalinguistic and sociopragmatic distinction |
| **HOUSE (2006)** | • M8,<br>• M9<br>• Generate | • Intercultural Pragmatics | • Theoretical | Five dimensions along which communicative norms vary |
| **LI & CAO (2019); LU (2019); MCGEE (2019)** | • M9. M4<br>• Generate | • Applied Linguistics | • Applied<br>• Review | Types and triggers of cross-cultural pragmatic failure |
| **HEIDER (1958)** | • M10<br>• Amplify | • Social Psychology | • Theoretical | Attribution of behaviour to disposition |
| **ROSS (1977)** | • M10<br>• Amplify | • Social Psychology | • Theoretical<br>• Experimental | Fundamental attribution error |
| **SCHEGLOFF, JEFFERSON, & SACKS (1977)** | • M11<br>• Detect | • Conversation Analysis | • Conversation-Analytic<br>• Theoretical | Repair as an organized system of self- and other-correction |
| **DINGEMANSE ET AL. (2015)** | • M11<br>• Detect | • Conversation Analysis<br>• Linguistic Typology | • Corpus (12 Languages) | Repair is universal and high-frequency, about once every 1.4 minutes |
| **RAYMOND & GILL (2025)** | • M11<br>• Detect | • Conversation Analysis | • Conversation-Analytic | Pre-emptive repair; its absence is a detect condition |
| **ALBERT & DE RUITER (2018)** | • M11<br>• Detect | • Conversation Analysis<br>• Cognitive Science | • Conversation-Analytic | Repair as a cross-disciplinary interface; source of the documented Case 1 |
| **SUCHMAN (1987)** | • M11<br>• Detect | • Human-Machine Interaction<br>• CSCW | • Analytic<br>• Ethnographic | False alarm and garden path defeat repair through unequal access to the situation |
| **CORTI & GILLESPIE (2016)** | • M11<br>• Detect | • Human-Machine Interaction<br>• Social Psychology | • Experimental | Repair initiation falls when the interlocutor is known to be an artificial agent, which is the experimental anchor for the claim that repair degrades under that condition |
| **PURVER, HOUGH, & HOWES (2018)** | • M11<br>• Detect (Operationalisation) | • Computational Linguistics | • Computational | Computational models cover repair and ambiguity-related miscommunication, providing a bridge to operationalisation and validation |

# Appendix B: Coding Manual

This appendix is an initial manual for applying the taxonomy to observed episodes. It is not itself a validation; it is the instrument a validation study would use. It has two parts: a general coding procedure, and a per-mechanism reference giving observable indicators, exclusion rules, and a test for distinguishing each mechanism from its nearest neighbor.

## B.1 General Procedure

Unit of analysis. A misunderstanding episode is a divergence between the meaning a speaker intended and the meaning a listener reconstructed, as defined in Section 3.1. The coder works from observable talk or interaction, not from assumed inner states.

Step 1: confirm it is a misunderstanding. Rule out the neighbouring phenomena in Section 3.2. A pure mishearing that produced no semantic divergence, an ambiguity in the signal that no one actually misread, a disagreement in which the meaning was understood and then rejected, and a successful deception are all out of scope. Include a case only when the listener reconstructed a meaning different from the intended one.

**Step 2: identify the misunderstood object.** Determine what diverged: message content, a referent, the speech act, an implied meaning, the intent or relational stance, or a requested future action. The object determines which mechanism is likely primary and which role a mechanism plays in this episode.

**Step 3: mark the primary mechanism.** The primary mechanism is the one that produced the divergence in the misunderstood object; it is the mechanism whose removal would have prevented the divergence. There is normally exactly one.

**Step 4: mark the contributing mechanisms.** A contributing mechanism enabled the divergence, made it worse, or hid it, but did not by itself produce it. There may be several, or none.

Step 5: record the role each mechanism played in the episode. Record whether each coded mechanism generated the divergence, amplified it, or governed its detection and repair. This may differ from the mechanism's dominant role, as Section 4.3 explains. Channel and Medium Distortion (M5) and Attributional Distortion (M10), in particular, may generate a divergence in some episodes.

Step 6: classify the episode. Record detection as undetected or detected and, if detected, resolution as resolved or unresolved, following Section 3.3.

## B.2 Per-Mechanism Reference

*Table 6. Coding Manual for Identifying and Distinguishing the Eleven Misunderstanding Mechanisms*

| MECHANISM | OBSERVABLE INDICATORS | DO NOT CODE WHEN | TEST AGAINST THE NEAREST NEIGHBOR |
|---|---|---|---|
| **1. INTENT UNDERSPECIFICATION** | Vague directives like "handle this"<br>Missing critical constraints<br>Heavy revision in follow-up messages<br>The speaker cannot say what was meant when asked | The intention was clear but the wording allowed a wrong reading (that is M4) | Did the speaker have a formed intention? If not, M1<br>If yes but poorly worded, M4 |
| **2. COMMON-GROUND OVERESTIMATION** | A party proceeds as if some knowledge is shared when it is not<br>There is no explicit grounding<br>The gap is broad rather than tied to one word | A specific trigger carries the assumption (M7), or the party held accurate beliefs but processed egocentrically (M3) | Is the error a false belief that background is shared? If yes, M2 |
| **3. PERSPECTIVE-TAKING FAILURE** | A party interprets or words the message from their own vantage point under load or time pressure, despite having, or being able to get, accurate knowledge of the other's position | The party genuinely believed the background was shared (M2) | Is it an error of belief (M2) or of using what one already knows (M3)? |
| **4. ENCODING AMBIGUITY** | Polysemy, syntactic or scope ambiguity, an unclear pronoun, unspecified scalar terms like "soon" or "some", undefined jargon<br>The listener chose a valid but unintended reading | The words allowed only one reading, or the divergence is about an implied meaning rather than the literal wording (M8) | Could a competent listener have read the exact words another valid way, and did they? If yes, M4 |
| **5. CHANNEL AND MEDIUM DISTORTION** | Cues needed for interpretation are absent or degraded by the medium: no tone or gesture, delay, a lost audio segment, a garbled transcription, a missing attachment, a cropped image | The exchange is face-to-face with full cues, or the difficulty is cultural convention rather than the channel (see Case 4) | Is a channel actually mediating and degrading the message? Code as amplify if it worsens another failure<br>Code as generate only if the medium changed the content received. |
| **6. REFERENTIAL MISALIGNMENT** | The same referring expression, a noun phrase, a pronoun, a word like "this" or "there", or a loose description, maps to different entities for the two parties | The expression is not referential, or both parties in fact reached the same referent | Is the divergence about which thing is meant (M6) rather than what is implied (M8) or which act is performed (M9)? |
| | A later turn reveals the mismatch | | |
| **7. PRESUPPOSITION MISMATCH** | A specific trigger, a factive verb, a definite description, a cleft, a particle, or a change-of-state verb, carries a background assumption the listener does not share<br>The listener accepts it without challenge or is left with an incomplete reading | The mismatch is a broad miscalibration of shared knowledge with no specific trigger (M2) | Can you point to the word or construction that smuggles in the assumption? If yes, M7<br>If it is diffuse, M2 |
| **8. PRAGMATIC INFERENCE FAILURE** | An implied meaning, an implicature, an indirect request or refusal, or a relevance implication, is not recovered<br>The literal content was understood but the intended point was missed | The failure is about the literal wording (M4), a specific triggered assumption (M7), or the type of act (M9) | Was the miss about what was implied beyond the words? If yes, M8 |
| **9. ILLOCUTIONARY FORCE MISMATCH** | The listener treats the act as a different type: a request as a question, a warning as a joke, a criticism as a suggestion, a refusal as a request | The content or the referent diverged rather than the act type | Was the mistake about what kind of act was performed? If yes, M9 |
| **10. ATTRIBUTIONAL DISTORTION** | A party explains the other's communicative behaviour by character or motive rather than situation<br>A relational judgement forms and repair is discouraged | The divergence is in message content produced by another mechanism, unless a dispositional attribution additionally forms on top of it | Did a dispositional attribution create or entrench the misunderstanding? Code as generate if the misunderstood object is relational stance or intent, as in Case 5<br>Otherwise code as amplify and record any effect on repair separately. |
| **11. REPAIR FAILURE** | No checking, clarifying, confirming, or correcting<br>The divergence never surfaces<br>A repair restores flow without resolving the underlying divergence<br>Pre-emptive clarification is absent | Repair promptly exposed and resolved the divergence. Code M11 only when repair is absent, delayed, ineffective, incomplete, or restores conversational progress without resolving the underlying divergence. | Would an effective repair move have exposed or resolved the divergence, and was it absent or ineffective? If yes, M11. |